\pdfoutput=1
\documentclass{article}

\usepackage{iclr2021_conference,times}

\usepackage{amsmath,amsfonts,bm}

\def\eqref#1{equation~\ref{#1}}

\def\1{\bm{1}}

\DeclareMathAlphabet{\mathsfit}{\encodingdefault}{\sfdefault}{m}{sl}
\SetMathAlphabet{\mathsfit}{bold}{\encodingdefault}{\sfdefault}{bx}{n}

\newcommand{\softmax}{\mathrm{softmax}}
\newcommand{\sigmoid}{\sigma}

\usepackage[utf8]{inputenc} 
\usepackage[T1]{fontenc}    
\usepackage{hyperref}       
\usepackage{url}            
\usepackage{booktabs}       
\usepackage{amsfonts}       
\usepackage{nicefrac}       
\usepackage{microtype}      
\usepackage{setspace}
\usepackage{graphicx}
\usepackage{tikz}
\usepackage{multirow}
\usepackage{amsmath}
\usepackage{bbm}
\usepackage{xcolor}
\usepackage{pifont}
\usepackage{enumitem}
\usepackage{sidecap}
\iclrfinalcopy
\newcommand{\bx}{\mathbf{x}}

\newcommand{\bp}{\mathbf{p}}
\newcommand{\bh}{\mathbf{h}}
\newcommand{\bz}{\mathbf{z}}

\newcommand{\bM}{\mathbf{M}}

\newcommand{\bmm}{\mathbf{m}}

\newcommand{\cL}{\mathcal{L}}
\newcommand{\tb}[1]{\textbf{#1}}

\DeclareMathOperator*{\RNN}{RNN}

\DeclareMathOperator*{\Ber}{Ber}

\newcommand{\yes}{\ding{51}}

\definecolor{red}{RGB}{255,73,92}
\definecolor{blue}{RGB}{37,110,255}
\definecolor{violet}{RGB}{70,35,122}
\definecolor{green}{RGB}{61,220,151}

\newcommand{\pie}[1]{%
\begin{tikzpicture}
 \draw (0,0) circle (1ex);\fill (1ex,0) arc (0:#1:1ex) -- (0,0) -- cycle;
\end{tikzpicture}%
}
\newcommand{\cm}{\pie{360}}%
\newcommand{\hx}{\pie{180}}%
\newcommand{\xm}{\pie{0}}%

\newcommand{\mc}[3]{\multicolumn{#1}{#2}{#3}}
\newcommand{\mr}[2]{\multirow{#1}{*}{#2}}%

\newcommand{\ourmodel}{contextual prototypical memory}
\newcommand{\ourmodelshort}{CPM}
\newcommand{\ourchar}{RoamingOmniglot}
\newcommand{\ourroom}{RoamingRooms}
\newcommand{\ourimg}{RoamingImageNet}
\newcommand{\OnlineMatchingNet}{O-MN}
\newcommand{\OnlineIMP}{O-IMP}
\newcommand{\OnlineProtoNet}{O-PN}

\title{
Wandering Within a World: \\Online Contextualized Few-Shot Learning
}

\author{
  Mengye Ren${}^{1,3}$ \quad Michael L. Iuzzolino${}^{2}$ \quad Michael C. Mozer${}^{2,4}$ \quad Richard S. Zemel${}^{1,3,5}$\\\\
  ${}^1$University of Toronto \quad
  ${}^2$Google Research \quad
  ${}^3$Vector Institute \quad \\
  ${}^4$University of Colorado, Boulder \quad
  ${}^5$CIFAR \\
  \texttt{\{mren,zemel\}@cs.toronto.edu}\\ \texttt{\{michael.iuzzolino,mozer\}@colorado.edu}
}

\begin{document}

\maketitle

\begin{abstract}
We aim to bridge the gap between typical human and machine-learning environments by extending the
standard framework of few-shot learning to an online, continual setting. In this setting, episodes
do not have separate training and testing phases, and instead models are evaluated online while
learning novel classes. As in the real world, where the presence of spatiotemporal context helps us
retrieve learned skills in the past, our online few-shot learning setting also features an
underlying context that changes throughout time. Object classes are correlated within a context and
inferring the correct context can lead to better performance. Building upon this setting, we propose
a new few-shot learning dataset based on large scale indoor imagery that mimics the visual
experience of an agent wandering within a world. Furthermore, we convert popular few-shot learning
approaches into online versions and we also propose a new \emph{\ourmodel{}} model that can make use
of spatiotemporal contextual information from the recent past. 
\footnote{ Our code and dataset are released at: \texttt{https://github.com/renmengye/oc-fewshot-public}}
\end{abstract}
\section{Introduction}
\vspace{-0.1in}
In machine learning, many paradigms exist for training and evaluating models: standard
train-then-evaluate, few-shot learning, incremental learning, continual learning, and so forth. None
of these paradigms well approximates the naturalistic conditions that humans and artificial agents
encounter as they wander within a physical environment. Consider, for example, learning and
remembering peoples' names in the course of daily life. We tend to see people in a given
environment---work, home, gym, etc. We tend to repeatedly revisit those environments, with different
environment base rates, nonuniform environment transition probabilities, and nonuniform base rates
of encountering a given person in a given environment. We need to recognize when we do not know a
person, and we need to learn to recognize them the next time we encounter them. We are not always
provided with a name, but we can learn in a semi-supervised manner. And every training trial is
itself an evaluation trial as we repeatedly use existing knowledge and acquire new knowledge. In
this article, we propose a novel paradigm, \emph{online contextualized few-shot learning}, that
approximates these naturalistic conditions, and we develop deep-learning architectures well suited
for this paradigm.

In traditional few-shot learning (FSL)~\citep{omniglot,matchingnet}, training is episodic. Within an
isolated episode, a set of new classes is introduced with a limited number of labeled examples per
class---the \textit{support}  set---followed by evaluation on an unlabeled \textit{query} set. While
this setup has inspired the development of a multitude of meta-learning algorithms which can be
trained to rapidly learn novel classes with a few labeled examples, the algorithms are focused
solely on the few classes introduced in the current episode; the classes learned are not carried
over to future episodes. Although incremental learning and continual learning
methods~\citep{icarl,rebalance} address the case where classes are carried over, the episodic
construction of these frameworks seems artificial: in our daily lives, we do not learn new objects
by grouping them with five other new objects, process them together, and then move on.

To break the rigid, artificial structure of continual and few-shot learning, we propose a new
continual few-shot learning setting where environments are revisited and the total number of novel
object classes increases over time. Crucially, model evaluation happens on each trial, very much
like the setup in online learning. When encountering a new class, the learning algorithm is expected
to indicate that the class is ``new,'' and it is then expected to recognize subsequent instances of
the class once a label has been provided.

When learning continually in such a dynamic environment, contextual information can guide learning
and remembering. Any structured sequence provides \emph{temporal context}: the instances encountered
recently are predictive of instances to be encountered next. In natural environments, \emph{spatial
context}---information in the current input weakly correlated with the occurrence of a particular
class---can be beneficial for retrieval as well. For example, we tend to see our  boss in an office
setting, not in a bedroom setting. Human memory retrieval benefits from both  spatial and temporal
context~\citep{Howard2017, foundationsmemory}. In our online few-shot learning setting, we provide
spatial context in the presentation of each instance and temporal structure to sequences, enabling
an agent to learn from both spatial and temporal context. Besides developing and experimenting on a
toy benchmark using handwritten characters~\citep{omniglot}, we also propose a new large-scale
benchmark for online contextualized few-shot learning derived from indoor panoramic
imagery~\citep{matterport}. In the toy benchmark, temporal context can be defined by the
co-occurrence of character classes. In the indoor environment, the context---temporal and
spatial---is a natural by-product as the agent wandering in between different rooms.

We propose a model that can exploit contextual information, called \emph{\ourmodel{}}
(\emph{\ourmodelshort{}}), which incorporates an RNN to encode contextual information and a separate
prototype memory to remember previously learned classes (see Figure~\ref{fig:mainmodel}). This model
obtains significant gains on few-shot classification performance compared to models that do not
retain a memory of the recent past. We compare to classic few-shot algorithms extended to an online
setting, and
\ourmodelshort{} consistently achieves the best performance.

\looseness=-1000
The main contributions of this paper are as follows. First, we define an \emph{online contextualized
few-shot learning (OC-FSL)} setting to mimic naturalistic human learning. Second, we build three
datasets: 1) {\it \ourchar{}} is based on handwritten characters from
Omniglot~\citep{omniglot}; 2) {\it \ourimg{}} is based on images from ImageNet~\citep{imagenet}; and 3) {\it \ourroom{}} is our new few-shot learning dataset
based on indoor imagery~\citep{matterport}, which resembles the visual experience of a wandering
agent. Third, we benchmark classic FSL methods and also explore our \ourmodelshort{} model, which
combines the strengths of RNNs for modeling temporal context and Prototypical Networks
\citep{protonet} for memory consolidation  and rapid learning.
\begin{figure}[t]
\vspace{-0.5in}
\centering
\includegraphics[width=\textwidth,trim={0.0cm 7.5cm 1.0cm 0},clip]{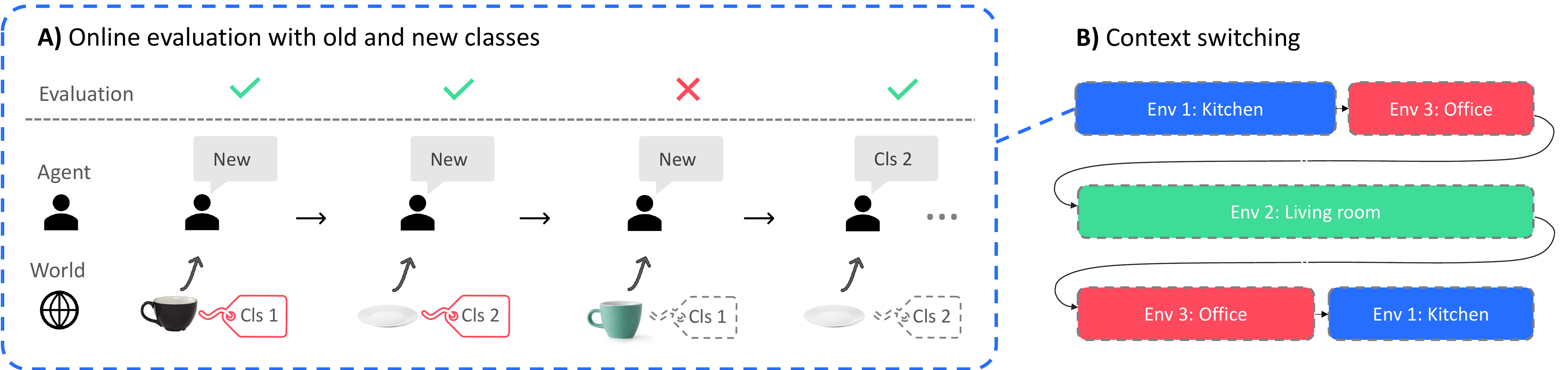}
\vspace{-0.3in}
\caption{\textbf{Online contextualized few-shot learning.} \textbf{A)} Our setup is similar to
online learning, where there is no separate testing phase; model training and evaluation happen
\textit{at the same time}. The input at each time step is an (image, class-label) pair. The number
of classes grows \textit{incrementally} and the agent is expected to answer ``new'' for items that
have not yet been assigned labels. Sequences can be \textit{semi-supervised}; here the label is not
revealed for every input item (labeled/unlabeled shown by red solid/grey dotted boxes). The agent is
evaluated on the correctness of all answers. The model obtains learning signals only on labeled
instances, and is correct if it predicts the label of previously-seen classes, or `new' for new
ones. \textbf{B)} The overall sequence switches between different \textit{learning environments}.
While the environment ID is \textit{hidden} from the agent, inferring the current environment can
help solve the task. }
\label{fig:setup}
\vspace{-0.15in}
\end{figure}
\begin{figure}[t]
\vspace{-0.5in}
\centering
\includegraphics[width=\textwidth, trim={0cm 7cm 6.5cm 0}, clip]{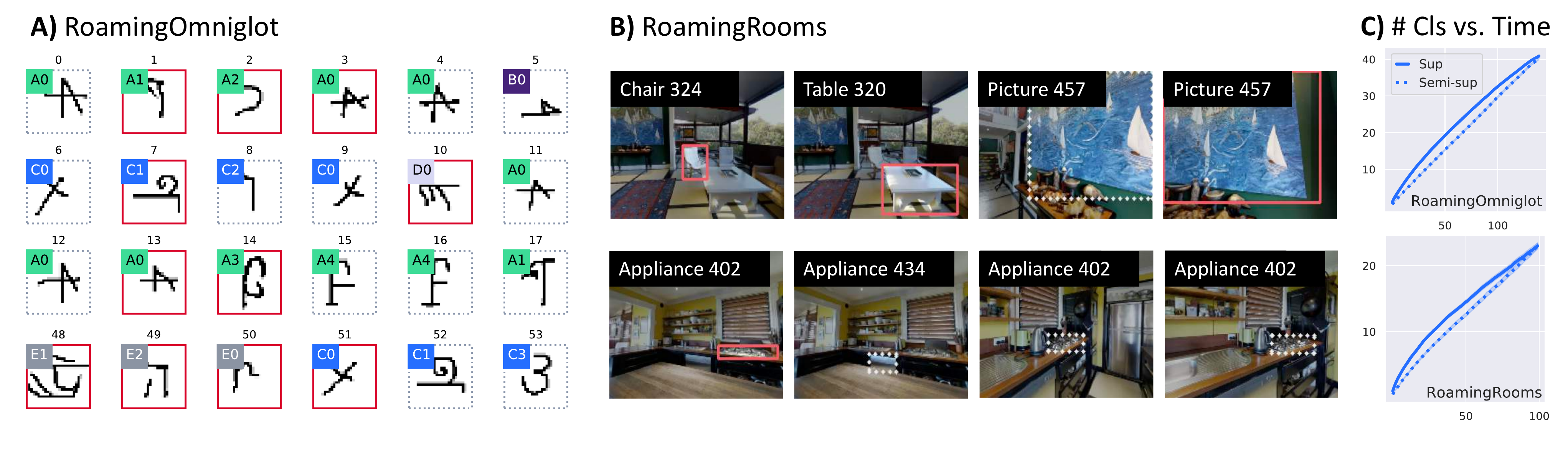}
\vspace{-0.32in}
\caption{\looseness=-1 \textbf{Sample online contextualized few-shot learning sequences.} \textbf{A)}
\ourchar{}. Red solid boxes denote labeled examples of Omniglot handwritten characters, and
dotted boxes denote unlabeled ones. Environments are shown in colored labels in the top left corner.
\textbf{B)} Image frame samples of a few-shot learning sequence in our \ourroom{} dataset collected
from a random walking agent. The task here is to recognize and classify novel instance IDs in the home environment. Here the groundtruth instance masks/bounding boxes are provided.
\textbf{C)} The growth of total number of labeled classes in a sequence for \ourchar{} (top) and
\ourroom{} (bottom).}
\label{fig:dataset}
\vspace{-0.15in}
\end{figure}

\section{Related Work}
\vspace{-0.1in}
In this section, we briefly review paradigms that have been used for few-shot learning (FSL) and
continual learning (CL). 

\vspace{-0.1in}
\paragraph{Few-shot learning:} 
FSL~\citep{omniglot,generativefewshot,siamese,matchingnet} considers learning new tasks with few
labeled examples. FSL models can be categorized as based on: metric
learning~\citep{matchingnet,protonet}, memory~\citep{mann}, and gradient
adaptation~\citep{maml,metasgd}. The model we propose, \ourmodelshort, lies on the boundary between
these approaches, as we use an RNN to model the temporal context but we also use metric-learning
mechanisms and objectives to train.

Several previous efforts have aimed to extend few-shot learning to incorporate more natural
constraints. One such example is semi-supervised FSL~\citep{fewshotssl}, where models also learn 
from a pool of unlabeled examples. While traditional FSL only
tests the learner on novel classes, \emph{incremental FSL}~\citep{lwof,attnattractor} tests on novel
classes together with a set of base classes. These approaches, however, have not explored how to
iteratively add new classes.

In concurrent work, \citet{contfsl} extend FSL to a continual setting based on image sequences, each
of which is divided into stages with a fixed number of examples per class followed by a query set. It
focuses on more flexible and faster adaptation since the models are evaluated online, and the context is
a soft constraint instead of a hard separation of tasks. Moreover, new classes need to be identified
as part of the sequence, crucial to any learner's incremental acquisition of knowledge.

\vspace{-0.1in}
\paragraph{Continual learning:} Continual (or lifelong) learning is a parallel line of research that
aims to handle a sequence of dynamic tasks~\citep{ewc,lwf,gem,expandable}. A key challenge here is
catastrophic forgetting~\citep{mccloskey1989catastrophic}, where the model ``forgets'' a task that
has been learned in the past. Incremental learning~\citep{icarl,eeil,bic,rebalance,inconline} is a
form of continual learning, where each task is an incremental batch of several new classes. This
assumption that novel classes always come in batches seems unnatural.

Traditionally, continual learning is studied with tasks such as permuted MNIST~\citep{mnist} or
split-CIFAR~\citep{cifar}. Recent datasets aim to consider more realistic continual learning, such as
CORe50~\citep{core50} and OpenLORIS~\citep{openloris}. We summarize core features of these continual
learning datasets in Appendix~\ref{app:data}. First, both CORe50 and OpenLORIS have relatively few object
classes, which makes meta-learning approaches inapplicable; second, both contain images of
small objects with minimal occlusion and viewpoint changes; and third, OpenLORIS does not have the
desired incremental class learning.

In concurrent work, \citet{osaka} proposes a setup to unify continual learning and meta-learning with
a similar online evaluation procedure. However, there are several notable differences. First, their
models focus on a general loss function without a specific design for predicting new classes; they
predict new tasks by examining if the loss of 
exceeds some threshold.
Second, the sequences of inputs are fully supervised. 
Lastly, their benchmarks are based on synthetic task sequences such as Omniglot or tiered-ImageNet, which are less naturalistic than our RoamingRooms dataset.

\vspace{-0.1in}
\paragraph{Online meta-learning:}
Some existing work builds on early approaches~\citep{thrun,Schmidhuber1987evolutionary} that tackle
continual learning from a meta-learning perspective. \citet{onlinemeta} proposes storing all task
data in a data buffer
; by contrast, \citet{oml} proposes to
instead learn a good representation that
supports such online updates. In \citet{onlinemixture}, a hierarchical Bayesian mixture model is used
to address the dynamic nature of continual learning. 

\vspace{-0.1in}
\paragraph{Connections to the human brain:}
Our \ourmodelshort\ model consists of multiple memory systems, consistent with claims of cognitive
neuroscientists of multiple memory systems in the brain. The  complementary learning systems (CLS)
theory~\citep{cls} suggests that the hippocampus stores the recent experience and is likely where
few-shot learning takes place. However, our model is more closely related to contextual binding
theory~\citep{ctxbinding}, which suggests that long-term encoding of information depends on binding
spatiotemporal context, and without this context as a cue, forgetting occurs. Our proposed
\ourmodelshort\ contains parallels to human brain memory components~\citep{cohensquire}. Long term
statistical learning is captured in a CNN that produces a deep embedding. An RNN holds a type of
working memory that can retain novel objects and spatiotemporal contexts. Lastly, the prototype
memory represents the semantic memory, which consolidates multiple events into a single knowledge
vector~\citep{semanticmem}. Other deep learning researchers have proposed multiple memory
systems for continual learning. In \citet{dualmem}, the learning algorithm is heuristic and
representations come from pretrained networks. In \citet{fearnet}, a prototype memory is used for
recalling recent examples and rehearsal from a generative model allows this knowledge to be
integrated and distilled into a long-term memory.

\section{Online Contextualized Few-Shot Learning}
\vspace{-0.1in}
\label{sec:benchmark}
In this section, we introduce our new online contextualized few-shot learning (OC-FSL) setup in the
form of a sequential decision problem, and then introduce our new benchmark datasets.

\vspace{-0.1in}
\paragraph{Continual few-shot classification as a sequential decision problem:}
In traditional few-shot learning, an episode is constructed by a support set $S$ and a query set
$Q$. A few-shot learner $f$ is expected to predict the class of each example in the query set
$\bx^Q$ based on the support set information: $\hat{y}^Q = f(\bx^Q; (\bx^S_1, y^S_1), \dots,
(\bx^S_N, y^S_N))$, where $\bx \in \mathbb{R}^D$, and $y \in [1 \dots K]$. This setup is not a natural fit for continual learning, since it is unclear when
to insert a query set into the sequence.

Inspired by the online learning literature, we can convert continual few-shot learning into a
sequential decision problem, where every input example is also part of the evaluation: $\hat{y}_t =
f(\bx_t; (\bx_1, \tilde{y}_1), \dots, (\bx_{t-1}, \tilde{y}_{t-1}))$, for $t = 1 \dots T$, where
$\tilde{y}$ here further allows that the sequence of inputs to be semi-supervised: $\tilde{y}$
equals $y_t$ if labeled, or otherwise $-1$. The setup in \citet{mann} and \citet{rareevent} is
similar; they train RNNs using such a temporal sequence as input. However, their evaluation relies
on a ``query set'' at the end. We instead evaluate online while learning.

Figure~\ref{fig:setup}-A illustrates these features, using an example of an input sequence where an
agent is learning about new objects in a kitchen. The model is rewarded when it correctly predicts a
known class or when it indicates that the item has yet to be assigned a label.

\vspace{-0.1in}
\paragraph{Contextualized environments:}
Typical continual learning consists of a sequence of tasks, and models are trained sequentially for
each task. This feature is also preserved in many incremental learning settings~\citep{icarl}. For
instance, the split-CIFAR task divides CIFAR-100 into 10 learning environments, each with 10
classes, presented sequentially. In our formulation, the sequence returns to earlier environments
(see Figure~\ref{fig:setup}-B), which enables assessment of long-term durability of knowledge.
Although the ground-truth environment identity is not provided, we structure the task such that the
environment itself provides contextual cues which can constrain the correct class label.
\emph{Spatial} cues in the input help distinguish one environment from another. \emph{Temporal} cues
are implicit because the sequence tends to switch environments infrequently, allowing recent inputs
to be beneficial in guiding the interpretation of the current input.

\begin{figure}[t]
\centering
\vspace{-0.5in}
\includegraphics[width=\textwidth,trim={0cm 12.5cm 2.5cm 0cm},clip]{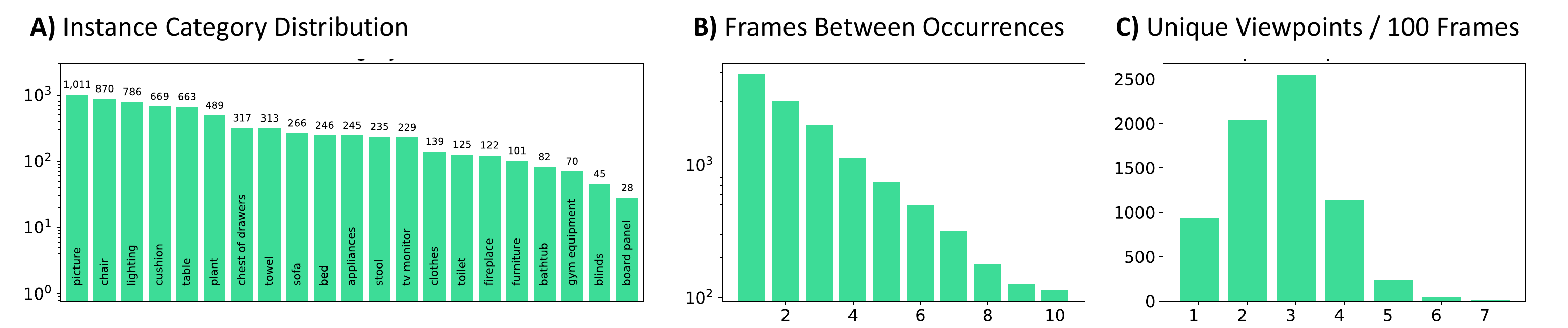}
\vspace{-.25in}
\caption{\textbf{Statistics for our \ourroom{} dataset.}
Plots show a natural long tail distribution of instances grouped into categories. An
average sequence has 3 different view points. Sequences are highly correlated in time but revisits
are not uncommon. }
\label{fig:dataset_distribution}
\vspace{-0.15in}
\end{figure}

\vspace{-0.1in}
\paragraph{\ourchar{}:}
The Omniglot dataset~\citep{omniglot} contains 1623 handwritten characters from 50 different
alphabets. We split the alphabets into 31 for training, 5 for validation, and 13 for testing. We
augment classes by 90 degree rotations to create 6492 classes in total. Each contextualized few-shot
learning image sequence contains 150 images, drawn from a random sample of 5-10 alphabets, for a
total of 50 classes per sequence. These classes are randomly assigned to 5 different environments;
within an environment, the characters are distributed according to a Chinese restaurant
process~\citep{crp} to mimic the imbalanced long-tail distribution of naturally occurring objects. We
switch between environments using a Markov switching process; i.e., at each step there is a constant
probability of switching to another environment. An example sequence is shown in
Figure~\ref{fig:dataset}-A.

\vspace{-0.1in}
\paragraph{\ourroom{}:}
As none of the current few-shot learning datasets provides the natural online learning experience
that we would like to study, we created our own dataset using simulated indoor environments. We
formulate this as a few-shot instance learning problem, which could be a use case for a home robot:
it needs to quickly recognize and differentiate novel object instances, and large viewpoint
variations can make this task challenging (see examples in Figure~\ref{fig:dataset}-B). There are
over 7,000 unique instance classes in the dataset, making it suitable to meta-learning approaches.

Our dataset is derived from the Matterport3D dataset~\citep{matterport} with 90 indoor worlds
captured using panoramic depth cameras. We split these into 60 worlds for training, 10 for
validation and 20 for testing. MatterSim~\citep{mattersim} is used to load the simulated world and
collect camera images and HabitatSim~\citep{habitat} is used to project instance
segmentation labels onto 2D image space. We created a random walking agent to collect the virtual
visual experience. For each viewpoint in the random walk, we randomly sample one object from
the image sensor and highlight it with the available instance segmentation, forming an input {\it
frame}. Each viewpoint provides environmental context---the other objects present in the
room with the highlighted object.

Figure~\ref{fig:dataset_distribution}-A shows the object instance distribution. We see strong
temporal correlation, as 30\% of the time the same instance appears in the next frame
(Figure~\ref{fig:dataset_distribution}-B), but there is also a significant proportion of revisits.
On average, there are three different viewpoints per 100-image sequence
(Figure~\ref{fig:dataset_distribution}-C). Details and other statistics of our proposed datasets are
included in the Appendix.

\vspace{-0.1in}
\paragraph{\ourimg{}:} To make the classification problem more challenging, we also report results 
on \ourimg, which uses the same online sequence sampler as \ourchar{} but applied on the \textit{tiered}-ImageNet 
dataset~\citep{fewshotssl}, a subset of the original ImageNet dataset~\citep{imagenet}.
It contains 608 classes with a total of 779K images of size 84$\times$84.
We use the same split of classes as \citet{fewshotssl}. The high-level categories are used for
sampling different ``environments'' just like the notion of alphabets in \ourchar{}.
\section{Contextual Prototypical Memory Networks}
\begin{figure}[t]
\vspace{-0.7in}
\begin{minipage}[c]{0.5\linewidth}
\centering
\vspace{0.25in}
\includegraphics[width=\linewidth,trim={0 8cm 4.5cm 0.3cm},clip]{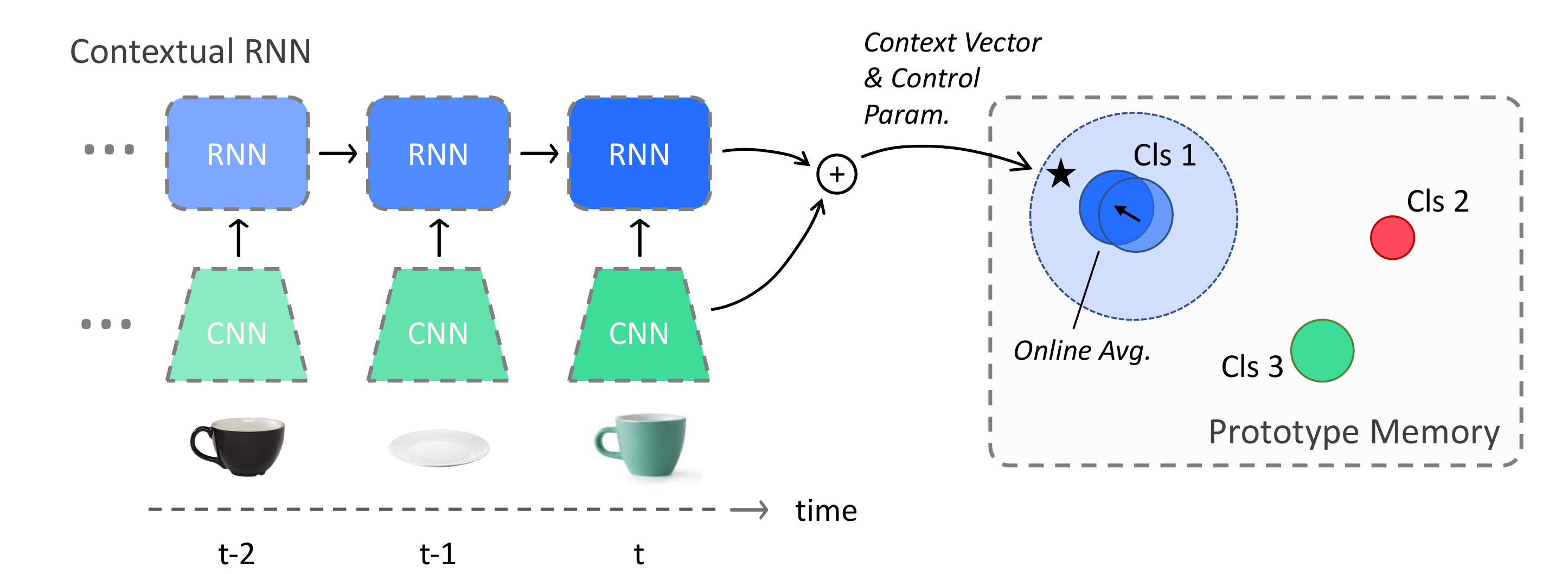}
\end{minipage}
\begin{minipage}[c]{0.5\linewidth}
\caption{\textbf{Contextual prototypical memory.} Temporal contextual features are extracted
from an RNN. The prototype memory stores one vector per class and does online averaging.
Examples falling outside the radii of all prototypes are classified as ``new.'' }
\label{fig:mainmodel}
\end{minipage}
\vspace{-0.25in}
\end{figure}
\vspace{-0.1in}
In the online contextualized few-shot learning setup, the few-shot learner can potentially improve
by modeling the temporal context. Metric learning approaches~\citep{protonet} typically ignore
temporal contextual relations and directly compare the similarity between training and test samples.
Gradient-based approaches~\citep{oml}, on the other hand, have the ability to adapt to new contexts,
but they do not naturally handle new and unlabeled examples. Inspired by the
contextual binding theory of human memory ~\citep{ctxbinding}, we propose a simple yet effective
approach that uses an RNN to transmit spatiotemporal context and control signals to a prototype memory
(Figure~\ref{fig:mainmodel}).


\vspace{-0.1in} \paragraph{Prototype memory:} We start describing our model with the prototype
memory, which is an online version of the Prototypical Network (or
\emph{ProtoNet})~\citep{protonet}. ProtoNet can be viewed as a knowledge base memory, where each
object class $k$ is represented by a prototype vector $\bp[k]$, computed as the mean vector of all
the support instances of the class in a sequence. It can also be applied to our task of online
few-shot learning naturally, with some modifications. Suppose that at time-step $t$ we have already
stored a few classes in the memory, each represented by their current prototype $\bp_t[k]$, and we
would like to query the memory using the input feature $\bh_t$. We model our prototype memory as %
$\hat{y}_{t,k} = \softmax( -d_{\bmm_t}(\bh_t, \bp_t[k]))$, 
(e.g. squared Euclidean distance or cosine dissimilarity) parameterized by a vector $\bmm_t$ that
scales each dimension with a Hadamard product. To predict whether an example is of a new class, we
can use a separate \emph{novelty} output $\hat{u}_t^{r}$ with sigmoid activation, similar to the
approach introduced in~\citet{fewshotssl}, where $\beta_t^r$ and $\gamma_t^r$ are
yet-to-be-specified thresholding hyperparameters (the superscript $r$ stands for read): 
\vskip -0.5cm 
\begin{align}
\hat{u}_t^{r} &= \sigmoid( (\min_k d_{\bmm_t}(\bh_t, \bp_t[k]) - \beta_t^r) /
\gamma_t^r).
\label{eq:ur}
\end{align}
\vskip -0.25cm

\vspace{-0.1in}
\paragraph{Memory consolidation with online prototype averaging:}
Traditionally, ProtoNet uses the average representation of a class across all support examples.
Here, we must be able to adapt the prototype memory incrementally at each step. 
Fortunately, we can recover the computation of a ProtoNet by performing a simple online averaging:
\vskip -0.8cm
\begin{align}
c_t &= c_{t-1} + 1; \ \ A(\bh_t; \bp_{t-1}, c_t) = \frac{1}{c_t} \left( \bp_{t-1} c_{t-1} +\bh_t \right),
\end{align}
\vskip -0.3cm
where $\bh$ is the input feature, and $\bp$ is the prototype, and $c$ is a scalar indicating the
number of examples that have been added to this prototype up to time $t$. The online averaging function
$A$ can also be made more flexible to allow more plasiticity, modeled by a
\textit{gated averaging unit} (GAU):
\vskip -0.5cm
\begin{align}
A_{\text{GAU}}(\bh_t; \bp_{t-1}) &= (1-f_t) \cdot \bp_{t-1} + f_t \cdot \bh_t, \ \ \ \text{where} \
\ f_t = \sigmoid(W_f [\bh_t, \bp_{t-1}] + b_f) \in \mathbb{R}.
\end{align}
\vskip -0.25cm
When the current example is
unlabeled, $\tilde{y}_t$ is encoded as $-1$, and the model's own prediction $\hat{y}_t$ will determine
which prototype to update; in this case, the model must also determine a strength of belief,
$\hat{u}_t^w$, that the current unlabeled example should be treated as a new class.  Given
$\hat{u}_t^w$ and $\hat{y}_t$, the model can then update a prototype:
\vskip -0.5cm
\begin{align}
\hat{u}_t^{w} &= \sigmoid((\min_k d_{\bmm_t}(\bh_t, \bp_t[k]) - \beta_t^w) / \gamma_t^w),\\
\Delta[k]_t   &= \underbrace{\mathbbm{1}[\tilde{y}_t = k]}_{\text{Supervised}} +
             \underbrace{\hat{y}_{t,k} (1 - \hat{u}_t^{w}) \mathbbm{1}[\tilde{y}_t =
-1]}_{\text{Unsupervised}}, \\ c[k]_t        &= c[k]_{t-1} + \Delta[k]_t,\\
\bp[k]_t      &= A(\bh_{t} \Delta[k]_t; \bp[k]_{t-1}, c[k]_t), \ \ \text{or} \ \ \bp[k]_t = A_{\text{GAU}}(\bh_{t} \Delta[k]_t; \bp[k]_{t-1}).
\end{align}
\looseness=-1
As-yet-unspecified hyperparameters $\beta_t^w$ and $\gamma_t^w$ are required (the superscript $w$
is for write). These parameters for the online-updating novelty
output $\hat{u}_t^{w}$ are  distinct from $\beta_t^r$ and $\gamma_t^r$ in Equation~\ref{eq:ur}. The
intuition is that for ``self-teaching'' to work, the model potentially needs to be more conservative
in creating new classes (avoiding corruption of prototypes) than in predicting an input as being a
new class.

\begin{table}[t]
\begin{center}
\vspace{-0.5in}
\caption{\textbf{\ourchar{} OC-FSL results.} Max 5 env, 150 images, 50 cls, with 8$\times$8 occlusion.}
\label{tab:omniglot}
\begin{small}
\resizebox{.75\textwidth}{!}{
\begin{tabular}{cccc|ccc}
\toprule
\mr{2}{\tb{Method}} & \mc{3}{c|}{\tb{Supervised}}                         &  \mc{3}{c}{\tb{Semi-supervised}}            \\
                     & AP         & 1-shot Acc.           & 3-shot Acc.           & AP         & 1-shot Acc.           & 3-shot Acc.           \\
\midrule         
LSTM                 & 64.34      & 61.00 $\pm$ 0.22      & 81.85 $\pm$ 0.21      & 54.34      & 68.30 $\pm$ 0.20      & 76.38 $\pm$ 0.49      \\
DNC                  & 81.30      & 78.87 $\pm$ 0.19      & 91.01 $\pm$ 0.15      & 81.37      & 88.56 $\pm$ 0.12      & 93.81 $\pm$ 0.26      \\
OML-U                & 77.38      & 70.98 $\pm$ 0.21      & 89.13 $\pm$ 0.16      & 66.70      & 74.65 $\pm$ 0.19      & 90.81 $\pm$ 0.34      \\
OML-U++              & 86.85      & 88.43 $\pm$ 0.14      & 92.01 $\pm$ 0.14      & 81.39      & 71.64 $\pm$ 0.19      & 93.72 $\pm$ 0.27      \\
\OnlineMatchingNet{} & 88.69      & 84.82 $\pm$ 0.15      & 95.55 $\pm$ 0.11      & 84.39      & 88.77 $\pm$ 0.13      & 97.28 $\pm$ 0.17      \\
\OnlineIMP{}         & 90.15      & 85.74 $\pm$ 0.15      & 96.66 $\pm$ 0.09      & 81.62      & 88.68 $\pm$ 0.13      & 97.09 $\pm$ 0.19      \\
\OnlineProtoNet{}    & 90.49      & 85.68 $\pm$ 0.15      & 96.95 $\pm$ 0.09      & 84.61      & 88.71 $\pm$ 0.13      & 97.61 $\pm$ 0.17      \\
CPM (Ours)           & \tb{94.17} & \tb{91.99} $\pm$ 0.11 & \tb{97.74} $\pm$ 0.08 & \tb{90.42}      & \tb{93.18} $\pm$ 0.16      & \tb{97.89} $\pm$ 0.15      \\
\bottomrule
\end{tabular}
}
\end{small}
\end{center}
\end{table}
\begin{table}[t]
\vspace{-0.15in}
\caption{\textbf{\ourroom{} OC-FSL results.} Max 100 images and 40 classes.}
\begin{center}
\begin{small}
\resizebox{.75\textwidth}{!}{
\begin{tabular}{cccc|ccc}
\toprule
\mr{2}{\tb{Method}}  & \mc{3}{c|}{\tb{Supervised}}                                &  \mc{3}{c}{\tb{Semi-supervised}}                          \\
                     & AP         & 1-shot Acc.           & 3-shot Acc.           & AP         & 1-shot Acc.           & 3-shot Acc.          \\
\midrule
LSTM                 & 45.67      & 59.90 $\pm$ 0.40      & 61.85 $\pm$ 0.45      & 33.32      & 52.71 $\pm$ 0.38      & 55.83 $\pm$ 0.76     \\
DNC                  & 80.86      & 82.15 $\pm$ 0.32      & 87.30 $\pm$ 0.30      & 73.49      & 80.27 $\pm$ 0.33      & 87.87 $\pm$ 0.49     \\
OML-U                & 76.27      & 73.91 $\pm$ 0.37      & 83.99 $\pm$ 0.33      & 63.40      & 70.67 $\pm$ 0.38      & 85.25 $\pm$ 0.56      \\
OML-U++              & 88.03      & 88.32 $\pm$ 0.27      & 89.61 $\pm$ 0.29      & 81.90      & 84.79 $\pm$ 0.31      & 89.80 $\pm$ 0.47      \\
\OnlineMatchingNet{} & 85.91      & 82.82 $\pm$ 0.32      & 89.99 $\pm$ 0.26      & 78.99      & 80.08 $\pm$ 0.34      & \tb{92.43} $\pm$ 0.41 \\
\OnlineIMP{}         & 87.33      & 85.28 $\pm$ 0.31      & 90.83 $\pm$ 0.25      & 75.36      & 84.57 $\pm$ 0.31      & 91.17 $\pm$ 0.43     \\
\OnlineProtoNet{}    & 86.01      & 84.89 $\pm$ 0.31      & 89.58 $\pm$ 0.28      & 76.36      & 80.67 $\pm$ 0.34      & 88.83 $\pm$ 0.49     \\
CPM (Ours)           & \tb{89.14} & \tb{88.39} $\pm$ 0.27 & \tb{91.31} $\pm$ 0.26 & \tb{84.12} & \tb{86.17} $\pm$ 0.30 & 91.16 $\pm$ 0.44     \\
\bottomrule
\end{tabular}
}
\label{tab:matterport}
\vspace{-0.25in}
\end{small}
\end{center}
\end{table}
\begin{table}[t]
\vspace{-0.5in}
\caption{\textbf{\ourimg{} OC-FSL results.} Max 150 images and 50 classes. * denotes CNN pretrained using regular classification.}
\begin{center}
\begin{small}
\resizebox{.75\textwidth}{!}{
\begin{tabular}{cccc|ccc}
\toprule
\mr{2}{\tb{Method}}  & \mc{3}{c|}{\tb{Supervised}}                                &  \mc{3}{c}{\tb{Semi-supervised}}                          \\
                     & AP         & 1-shot Acc.           & 3-shot Acc.           & AP         & 1-shot Acc.           & 3-shot Acc.          \\
\midrule
LSTM*                & 22.54      & 28.14 $\pm$ 0.20      & 52.07 $\pm$ 0.27      & 13.50      & 30.02 $\pm$ 0.20      & 46.95 $\pm$ 0.56     \\
DNC*                 & 26.80      & 33.45 $\pm$ 0.19      & 55.78 $\pm$ 0.27      & 16.50      & 39.53 $\pm$ 0.19      & 54.10 $\pm$ 0.54     \\
OML-U                & 21.89      & 15.06 $\pm$ 0.14      & 52.52 $\pm$ 0.27      & 10.16      & 22.74 $\pm$ 0.17      & 55.81 $\pm$ 0.55     \\
\OnlineMatchingNet{} & 13.05      & 20.61 $\pm$ 0.15      & 38.73 $\pm$ 0.24      & 9.32       & 25.96 $\pm$ 0.16      & 55.32 $\pm$ 0.51     \\
\OnlineIMP{}         & 14.25      & 22.92 $\pm$ 0.16      & 41.01 $\pm$ 0.25      & 4.55       & 20.70 $\pm$ 0.15      & 51.23 $\pm$ 0.53     \\
\OnlineProtoNet{}*   & 23.10      & 32.82 $\pm$ 0.19      & 49.98 $\pm$ 0.25      & 15.76      & 36.69 $\pm$ 0.18      & 55.47 $\pm$ 0.53     \\
CPM (Ours)           & \tb{34.43} & \tb{40.40} $\pm$ 0.21 & \tb{60.29} $\pm$ 0.26 & \tb{24.75} & \tb{44.58} $\pm$ 0.21 & \tb{58.72} $\pm$ 0.53\\
\bottomrule
\end{tabular}
}
\vspace{-0.2in}
\label{tab:imagenet}
\end{small}
\end{center}
\end{table}

\vspace{-0.1in}
\paragraph{Contextual RNN:}
Instead of directly using the features from the CNN $\bh_t^{\text{CNN}}$ as input features to the
prototype memory, we would also like to use contextual information from the recent past. Above we
introduced threshold hyperparameters $\beta_t^r$, $\gamma_t^r$, $\beta_t^w$, $\gamma_t^w$ as well as
the metric parameter $\bM_t$. We let the contextual RNN output these additional control parameters,
so that the unknown thresholds and metric function can adapt based on the information in the
context. The RNN produces the context vector $\bh_t^{\text{RNN}}$ and other control parameters
conditioned on  $\bh_t^{\text{CNN}}$:
\begin{align}
\left[ \bz_t, \bh_t^{\text{RNN}}, \bmm_t, \ \beta^r_t, \ \gamma^r_t, \ \beta^w_t, \ \gamma^w_t
\right] =
\RNN(\bh_t^{\text{CNN}}; \bz_{t-1}),
\end{align}
where $\bz_t$ is the recurrent state of the RNN, and $\bmm_t$ is the scaling factor in the
dissimilarity score. The context, $\bh_t^{\text{RNN}}$, serves as an additive bias on the state
vector used for FSL: $\bh_t  = \bh_t^{\text{CNN}} + \bh_t^{\text{RNN}}$. This addition operation in
the feature space can help contextualize prototypes based on temporal proximity, and is also similar
to how the human brain leverages spatiotemporal context for memory storage~\citep{ctxbinding}.

\vspace{-0.1in}
\paragraph{Loss function:}
The loss function is computed after an entire sequence ends and all network parameters are learned
end-to-end. The loss is composed of two parts. The first is binary cross-entropy (BCE), for telling
whether each example has been assigned a label or not, i.e., prediction of new classes, and $u_t$ is the ground-truth binary label. Second we
use a multi-class cross-entropy for classifying among the known ones. We can write down the overall
loss function as follows:
\vskip -0.5cm
\begin{align}
\cL = \frac{1}{T} \sum_{t=1}^{T}
  \lambda \underbrace{ \left[ 
- u_t \log(\hat{u}_t^r) 
- (1-u_t) \log(1 - \hat{u}_t^r)  
\right]}_{\text{Binary cross entropy on old vs. new}}
+ \sum_{k=1}^{K} \underbrace{-\mathbbm{1}[y_t = k] (1-u_t) \log(\hat{y}_{t,k})}_{\text{Cross entropy on old classes}}
.
\label{eq:loss}
\end{align}

\vspace{-0.1in}
\paragraph{Training and evaluation:} During training, we sample learning sequences and for each sequence, we perform one iterative update to minimize the loss function (Eq.~\ref{eq:loss}). At the beginning of each sequence, the memory is reset. During training, the model learns from a set of training classes. During test time, the model recognizes new classes that have never been seen during training.
\section{Experiments}
\vspace{-0.1in}
In this section, we show experimental results for our online contextualized few-shot learning
paradigm, using \ourchar{} and \ourroom{} (see Sec.~\ref{sec:benchmark}) to evaluate our model, CPM,
and other state-of-the-art methods. For Omniglot, we apply an 8$\times$8 CutOut~\citep{cutout} to
each image to make the task more challenging. 

\vspace{-0.1in} \paragraph{Implementation details:} For \ourchar{}, we use the common
4-layer CNN for few-shot learning with 64 channels in each layer. For \ourimg{}, we also use ResNet-12 
with input resolution 84$\times$84~\citep{tadam}. For the \ourroom{}, we resize the input to 
120$\times$160 and use ResNet-12. To represent
the feature of the input image with an attention mask, we concatenate the global average pooled
feature with the attention ROI feature, resulting in a 512d feature vector. For the contextual RNN,
in both experiments we used an LSTM~\citep{lstm} with a 256d hidden state. The best CPM model is
equipped using GAU and cosine similarity for querying prototypes. Logits based on cosine similarity
are multiplied with a learned scalar initialized at 10.0~\citep{tadam}. We include additional
training details in Appendix~\ref{app:exp}.

\vspace{-0.1in}
\paragraph{Evaluation metrics:}
In order to compute a single number that characterizes the learning ability over sequences, we
propose to use \textit{average precision} (AP) to evaluate
both with respect to old versus new and the
specific class predictions.
Concretely, all predictions are sorted by their old vs. new scores, and we compute AP using the area
under the precision-recall curve. A true positive is defined as the correct prediction of a
multi-class classification among known classes. We also compute the ``$N$-shot'' accuracy; i.e., the
average accuracy after seeing the label $N$ times in the sequence. Note that these accuracy scores
only reflect the performance on {\it known} class predictions. All numbers are reported with an
average over 2,000 sequences and for $N$-shot accuracy standard error is also included.
Further explanation of these metrics is in Appendix~\ref{sec:metrics}. 

\vspace{-0.1in}
\paragraph{Comparisons:}
To evaluate the merits of our proposed model, we implement classic few-shot learning and online
meta-learning methods. More implementation and training details of these baseline methods can be
found in Appendix~\ref{app:exp}.
\begin{figure}[t]
\vspace{-0.5in}
\centering
\setlength{\tabcolsep}{0pt}
\begin{tabular}{cccc}
\includegraphics[height=2.4cm,trim={0.3cm 0cm 0.5cm 0},clip]{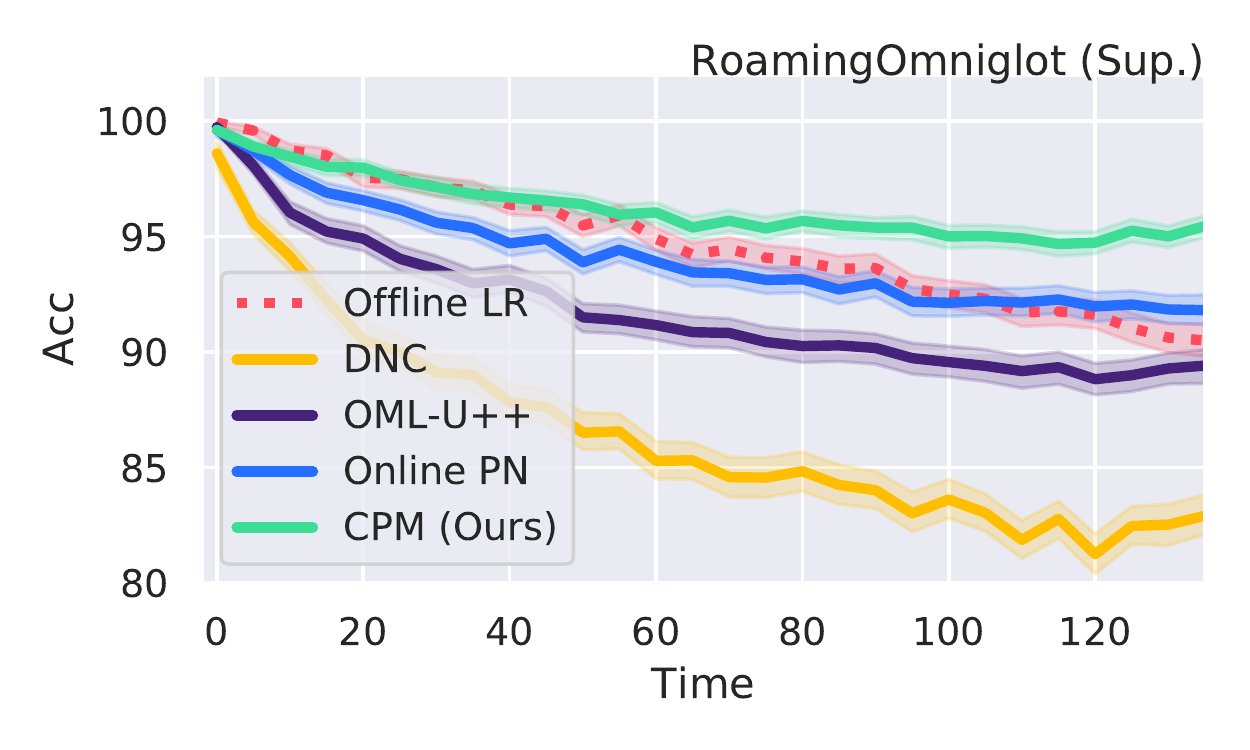}
&
\includegraphics[height=2.4cm,trim={2cm 0cm 0cm 0},clip]{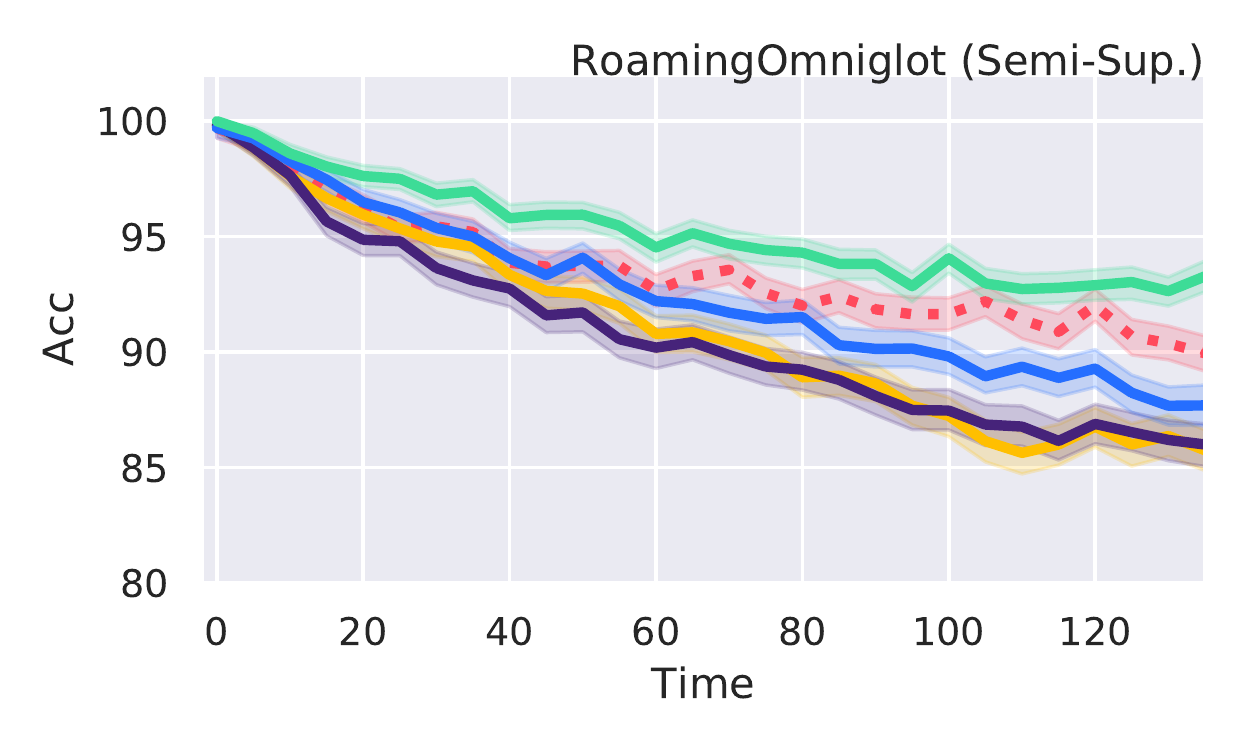}
&
\includegraphics[height=2.4cm,trim={1cm 0cm 0.5cm 0},clip]{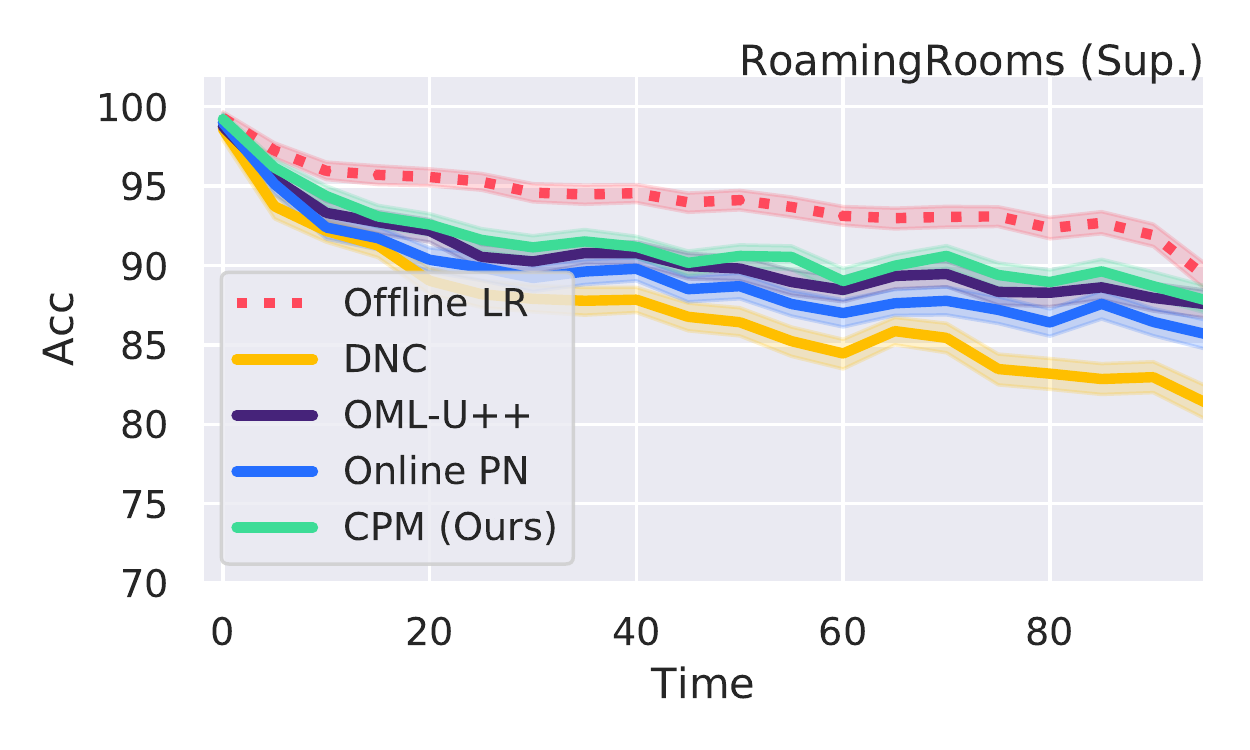}
&
\includegraphics[height=2.4cm,trim={2cm 0cm 0cm 0},clip]{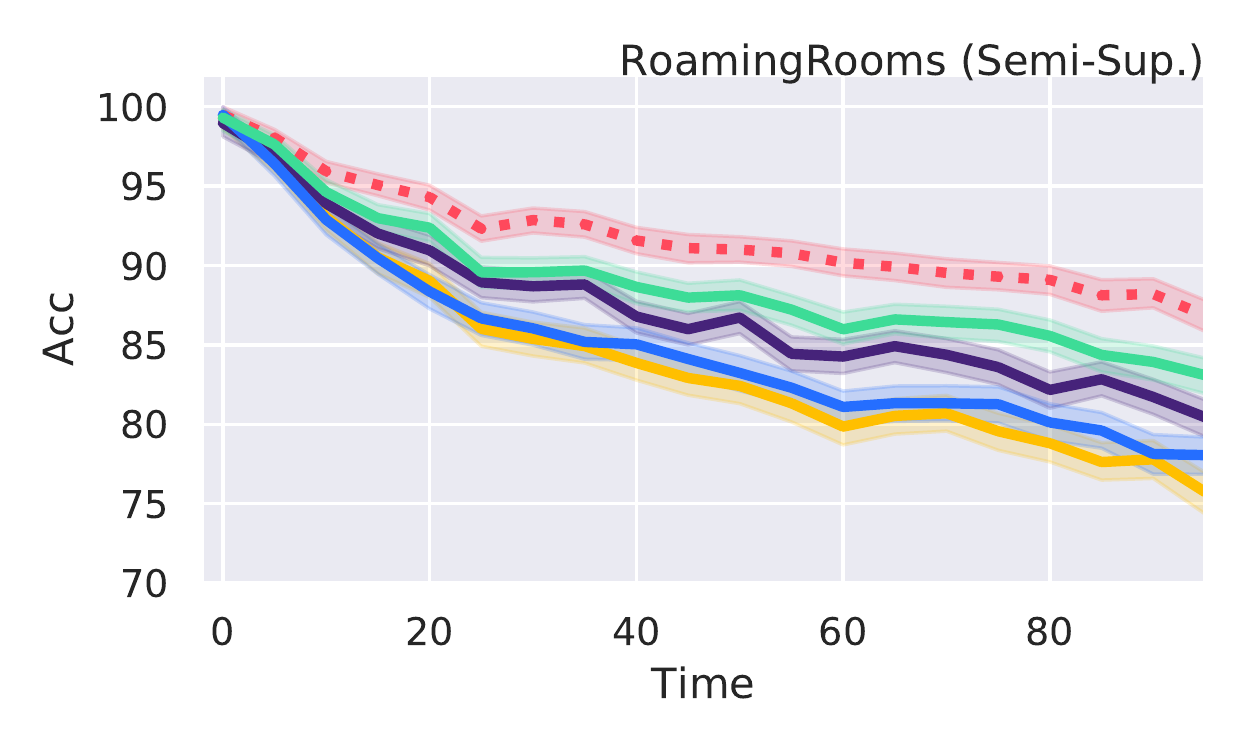}
\\
\end{tabular}
\vspace{-0.25in}
\caption{\textbf{Few-shot classification accuracy over time.} \textbf{Left:} \ourchar{}.
\textbf{Right:} \ourroom{}. \textbf{Top:} Supervised. \textbf{Bottom:} Semi-supervised. An offline
logistic regression (Offline LR) baseline is also included, using pretrained ProtoNet features. It
is trained on all labeled examples except for the one at the current time step.}
\label{fig:acctimefull}
\end{figure}
\begin{figure}[t]
\vspace{-0.1in}
\centering
\includegraphics[height=2.8cm,trim={-2.25cm 10cm 20cm 0.5cm},clip]{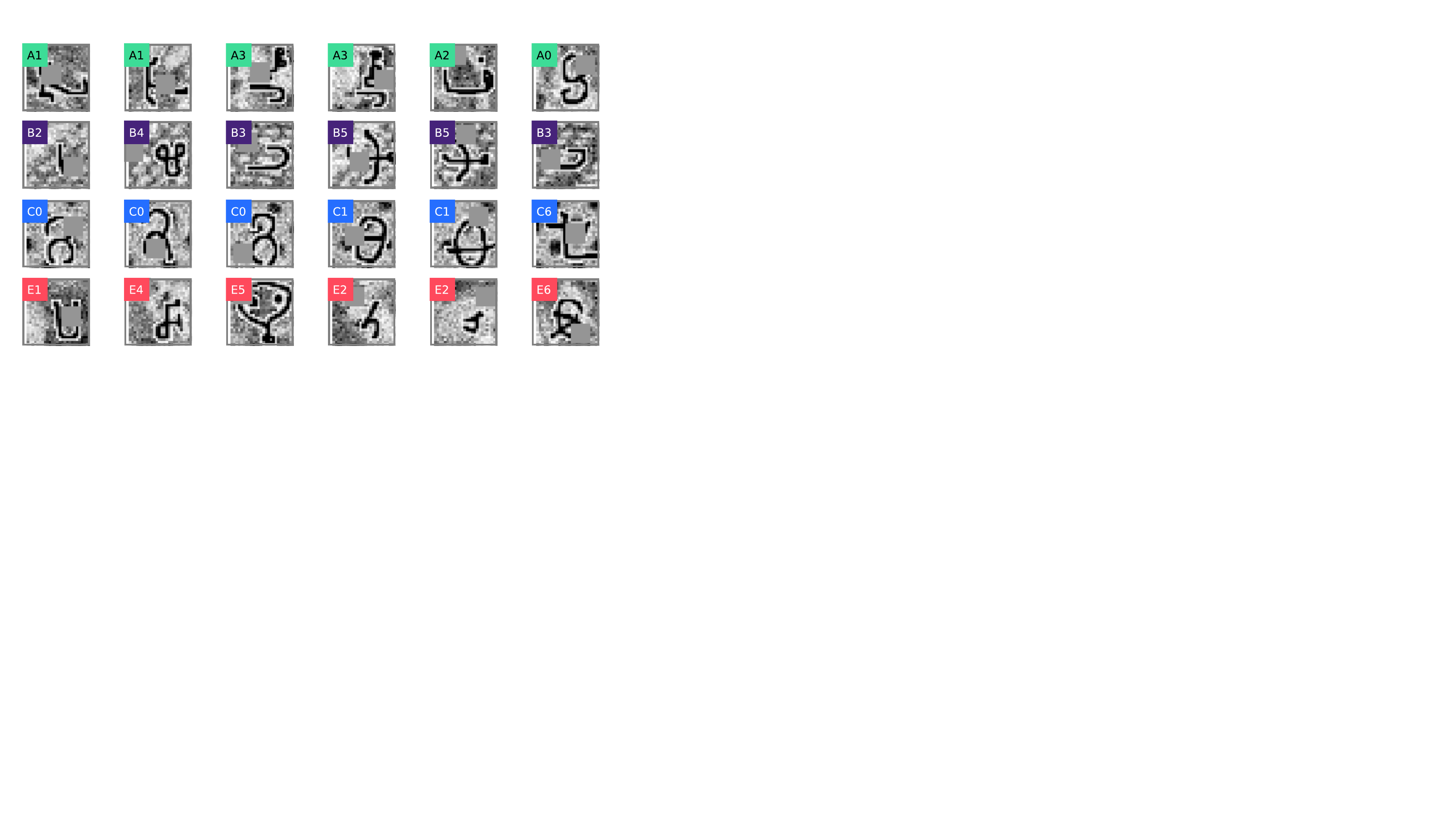}
\includegraphics[height=2.8cm,trim={-2.25cm 0 0.5cm 0},clip]{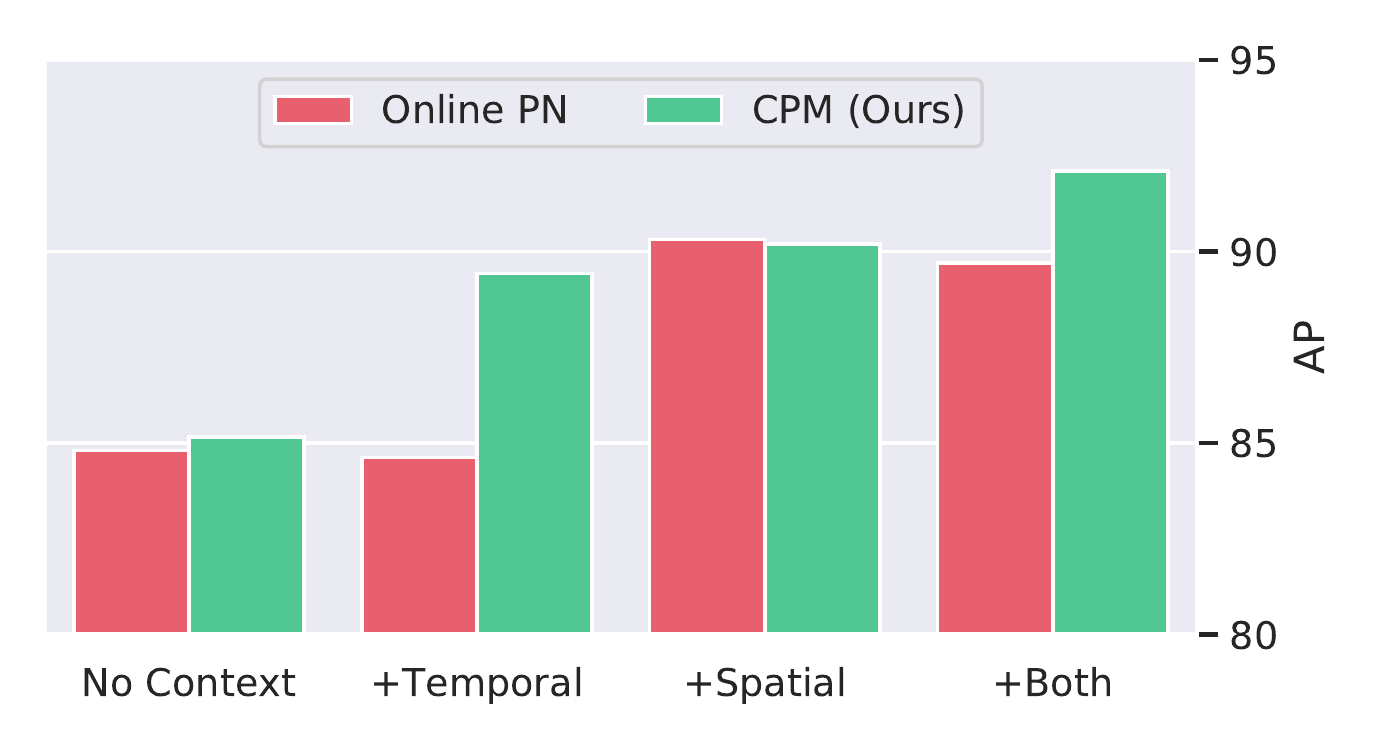}
\quad
\vspace{-0.2in}
\caption{\textbf{Effect of spatiotemporal context.} Spatiotemporal
context are added separately and together in \ourchar{}, by introducing texture background and
temporal correlation. \textbf{Left:} Stimuli used for spatial cue of the background environment.
\textbf{Right:} Our CPM model benefits from the presence of a temporal context
(``+Temporal'' and ``+Both'')} 
\label{fig:spatiotemporal}
\end{figure}
\vspace{-0.1in}
\begin{itemize}[leftmargin=*]
    \item \textbf{OML}~\citep{oml}: This is an online version of MAML~\citep{maml}. It performs one
    gradient descent step for each labeled input image, and slow weights are learned via
    backpropagation through time. On top of OML, we added an unknown predictor $\hat{u}_t = 1 -
    \max_k \hat{y}_{t,k}$ \footnote{We tried a few other ways and this is found to be the
    best.} (\textbf{OML-U}). We also found that using cosine classifier without the last layer ReLU
    is usually better than using the original dot-product classifier, and this improvement is denoted
    as \textbf{OML-U++}.
    \item \textbf{LSTM}~\citep{lstm} \& \textbf{DNC}~\citep{dnc}: We include RNN methods for
    comparison as well. Differentiable neural computer (DNC) is an improved version of memory
    augmented neural network (MANN)~\citep{mann}.
    \item \textbf{Online MatchingNet (\OnlineMatchingNet{})}~\citep{matchingnet}, 
    \textbf{IMP (\OnlineIMP{})}~\citep{imp} \&
    \textbf{ProtoNet (\OnlineProtoNet{})}~\citep{protonet}: We used\ the same negative Euclidean distance as the
    similarity function for these three metric learning based approaches.  In particular,
    MatchingNet stores all examples and performs nearest neighbor matching, which can be memory
    inefficient. Note that Online ProtoNet is a variant of our method without the contextual RNN.
\end{itemize}
\begin{table}[t]
\vspace{-0.5in}
    \centering
    \caption{\textbf{Effect of forgetting over a time interval on \ourchar{}.} Average accuracy vs. the number of time steps since the model has last seen the label of a particular class.}
    \vspace{0.1in}
    \resizebox{\textwidth}{!}{
    \begin{tabular}{ccccccc|cccccc}
    \toprule
    & \mc{6}{c}{\bf Supervised} & \mc{6}{c}{\bf Semi-Supervised}\\
    Interval  &   1 - 2&      3 - 5&      6 - 10&     11 - 20&    21 - 50&    51 - 100&
                    1 - 2&      3 - 5&      6 - 10&     11 - 20&    21 - 50&    51 - 100\\
    \midrule
    OPN 1-Shot  &   88.8&       86.9&       85.2&       84.7&	    83.6&	    81.1&
                    90.1&       88.9&	    88.4&	    87.6&	    87.3&	    85.1\\
    CPM 1-Shot  &   \bf{96.1}&  \bf{94.0}&	\bf{93.0}&	\bf{91.6}&	\bf{88.2}&  \bf{84.6}&
                    \bf{95.9}&  \bf{93.8}&  \bf{92.8}&	\bf{91.8}&	\bf{89.4}&	\bf{85.7}\\
    \midrule
    OPN 3-Shot  &   97.2&	    97.1&	    96.6&	    96.7&	    \bf{96.5}&	95.3&
                    97.8&	    97.3&	    97.1&	    \bf{97.8}&	\bf{97.7}&  \bf{96.8}\\
    CPM 3-Shot  &   \bf{98.5}&	\bf{98.2}&	\bf{97.5}&	\bf{97.2}&      95.4&  \bf{95.5}&
                    \bf{98.7}&	\bf{97.5}&	\bf{97.5}&	    96.5&	    96.3&	    92.9\\
    \bottomrule
    \end{tabular}}
    \label{tab:forgetomniglot}
\end{table}

\newcommand{\bgstar}{\beta_t^*, \gamma_t^*}
\newcommand{\bgw}{\beta_t^w$, $\gamma_t^w}
\newcommand{\hrnn}{\bh^{\text{RNN}}}

\begin{table}[t]
\vspace{-0.1in}
\begin{minipage}[t]{0.45\textwidth}
\begin{small}
\caption{Ablation of CPM architectural components on \ourchar{}}
\begin{center}
\label{tab:ablation}
\resizebox{!}{1.4cm}{
\begin{tabular}{ccccccc}
\toprule
\tb{Method}                   & $\hrnn$ & $\bgstar$ & Metric $\bmm_t$ & GAU  & Val AP      \\
\midrule                                                            
\OnlineProtoNet{}             &         &           &                 &      & 91.22       \\
No $\bh^{\text{RNN}}$         &         &  \yes     & \yes            &      & 92.52       \\
$\bh^{\text{RNN}}$ only       & \yes    &           &                 &      & 93.48       \\
No metric $\bmm_t$             & \yes   &  \yes     &                 &      & 93.61       \\
No $\beta_t^*, \gamma_t^*$    & \yes    &           & \yes            &      & 93.98       \\
$\bh_t = \bh_t^{\text{RNN}}$  & \yes    &  \yes     & \yes            &      & 93.70       \\
CPM Avg. Euc               & \yes    &  \yes     & \yes            &      & 94.08       \\
CPM Avg. Cos               & \yes    &  \yes     & \yes            &      & 94.57       \\
CPM GAU Euc                  & \yes    &  \yes     & \yes            & \yes & 94.11       \\
CPM GAU Cos                  & \yes    &  \yes     & \yes            & \yes & \tb{94.65}  \\
\bottomrule
\end{tabular}
}
\end{center}
\end{small}
\end{minipage}
\hfill
\begin{minipage}[t]{0.45\textwidth}
\caption{Ablation of semi-supervised learning components on \ourchar{}}
\begin{small}
\begin{center}
\label{tab:ablationssl}
\resizebox{!}{1.4cm}{
\begin{tabular}{cccccccc}
\toprule
\tb{Method}         & RNN   & Prototype & $\bgw$ & GAU  & Val AP     \\
\midrule                                                                                   
\OnlineProtoNet{}   &       &           &        &      & 90.83      \\
\OnlineProtoNet{}   &       &  \yes     &        &      & 89.10      \\
\OnlineProtoNet{}   &       &  \yes     & \yes   &      & 91.22      \\
CPM                 &       &           &        &      & 92.57      \\
CPM                 &  \yes &           &        &      & 93.16      \\
CPM                 &  \yes &  \yes     &        &      & 93.20      \\
CPM                 &  \yes &  \yes     & \yes   &      & 94.08      \\
CPM                 &  \yes &  \yes     & \yes   & \yes & \tb{94.65} \\
\bottomrule
\end{tabular}
}
\end{center}
\end{small}
\end{minipage}
\vspace{-0.1in}
\end{table}

\vspace{-0.1in}
\paragraph{Main results:} Our main results are shown in Table~\ref{tab:omniglot}, 
\ref{tab:matterport} and \ref{tab:imagenet}, including both supervised and semi-supervised settings. Our approach achieves
the best performance on AP consistently across all settings. Online ProtoNet is a direct comparison
without our contextual RNN and it is clear that CPM is significantly better. Our method is slightly
worse than Online MatchingNet in terms of 3-shot accuracy on the \ourroom{} semisupervised
benchmark. This can be explained by the fact that MatchingNet stores all past seen examples, whereas
CPM only stores one prototype per class. Per timestep accuracy is plotted in
Figure~\ref{fig:acctimefull}, and the decaying accuracy is due to the increasing number of classes
over time. In \ourchar{}, CPM is able to closely match or even sometimes surpass the offline
classifier, which re-trains at each step and uses all images in a sequence except the current one. This is
reasonable as our model is able to leverage information from the current context.

\vspace{-0.1in}
\paragraph{Effect of spatiotemporal context:} To answer the question whether the gain in performance
is due to spatiotemporal reasoning, we conduct the following experiment comparing CPM with online
ProtoNet. We allow the CNN to have the ability to recognize the context in \ourchar{} by adding a
texture background image using the Kylberg texture dataset~\citep{uppsala} (see
Figure~\ref{fig:spatiotemporal} left). As a control, we can also destroy the temporal context by
shuffling all the images in a sequence. We train four different models on dataset controls with or
without the presence of spatial or temporal context, and results are shown in
Figure~\ref{fig:spatiotemporal}. First, both online ProtoNet and CPM benefit from the inclusion of a
spatial context. This is understandable as the CNN has the ability to learn spatial cues, which
re-confirms our  main hypothesis that successful inference of the current context is beneficial to
novel object recognition. Second, only our CPM model benefits from the presence of temporal context,
and it receives distinct gains from spatial and temporal contexts.

\vspace{-0.1in}
\paragraph{Effect of forgetting:} As the number of learned classes increases, we expect the average accuracy to drop. To further investigate this forgetting effect, we measure the average accuracy in terms of the number of time steps the model has last seen the label of a particular class. It is reported in Table~\ref{tab:forgetomniglot} and in Appendix~\ref{sec:additionalresults} Table~\ref{tab:forgetroom}, \ref{tab:forgetimagenet}, where we directly compare CPM and OPN to see the effect of temporal context.
CPM is significantly better than OPN on 1-shot within a short interval, which suggests that the contextual RNN 
makes the recall of the recent past much easier. On \ourimg{}, OPN eventually surpasses CPM on longer horizon, and this can be explained by the fact that OPN has more stable prototypes, whereas prototypes in CPM could potentially be affected by the fluctuation of the contextual RNN over a longer horizon.

\vspace{-0.1in}
\paragraph{Ablation studies:} We ablate each individual module we introduce. Results are shown in
Tables~\ref{tab:ablation} and~\ref{tab:ablationssl}. Table~\ref{tab:ablation} studies different ways
we use the RNN, including the context vector $\bh^{\text{RNN}}$, the predicted threshold parameters
$\beta_t^*,\gamma_t^*$, and the predicted metric scaling vector $\bm_t$. Table~\ref{tab:ablationssl}
studies various ways to learn from unlabeled examples, where we separately disable the RNN update,
prototype update, and distinct write-threshold parameters $\beta^w_t, \gamma^w_t$ (vs. using
read-threshold parameters), which makes it robust to potential mistakes made in
semi-supervised learning. We verify that each component has a positive impact on the performance.
\vspace{-0.1in}
\section{Conclusion}
\vspace{-0.1in}
We proposed online contextualized few-shot learning, OC-FSL, a paradigm for machine learning that
emulates a human or artificial agent interacting with a physical world. It combines multiple
properties to create a challenging learning task: every input must be classified or flagged as
novel, every input is also used for training, semi-supervised learning can potentially improve
performance, and the temporal distribution of inputs is non-IID and comes from a generative model in
which  input and class distributions are conditional on a latent environment with Markovian
transition probabilities. We proposed the \ourroom{} dataset to simulate an agent wandering within a
physical world. We also proposed a new model, CPM, which uses an RNN to extract spatiotemporal
context from the input stream and to provide control settings to a prototype-based FSL model. In the
context of naturalistic domains like \ourroom{}, CPM is able to leverage contextual information to
attain performance unmatched by other state-of-the-art FSL methods.
\paragraph{Acknowledgments:} 
We thank Fei Sha, James Lucas, Eleni Triantafillou and Tyler Scott for helpful feedback on an earlier draft of
the manuscript. 
Resources used in preparing this research were provided, in part, by the
Province of Ontario, the Government of Canada through CIFAR, and companies
sponsoring the Vector Institute (\url{www.vectorinstitute.ai/\#partners}).
This project is supported by NSERC and the Intelligence Advanced Research Projects
Activity (IARPA) via Department of Interior/Interior Business Center (DoI/IBC) contract number
D16PC00003. The U.S. Government is authorized to reproduce and distribute reprints for Governmental
purposes notwithstanding any copyright annotation thereon. Disclaimer: The views and conclusions
contained herein are those of the authors and should not be interpreted as necessarily representing
the official policies or endorsements, either expressed or implied, of IARPA, DoI/IBC, or the U.S.
Government.
{
\bibliography{ref}
\bibliographystyle{iclr2021_conference}
}
\setstretch{1.0}

\newpage
\appendix
\section{Dataset Details}
\label{app:data}

\subsection{Benchmark comparison}
We include Table~\ref{tab:benchmark} to compare existing continual and few-shot learning paradigms.

\subsection{\ourchar{} \& \ourimg{} Sampler Details}
For the \ourchar{} and the \ourimg{} experiments, we use sequences with maximum 150 images, from 5 environments. For
individual environment, we use a Chinese restaurant process to sample the class distribution. In
particular, the probability of sampling a new class is:
\begin{align}
p_\text{new} = \frac{k \alpha + \theta}{m + \theta},
\end{align}
where $k$ is the number of classes that we have already sampled in the environment, and $m$ is the
total number of instances we have in the environment. $\alpha$ is set to 0.2 and $\theta$ is set to
1.0 in all experiments.

The environment switching is implemented by a Markov switching process. At each step in the
sequence there is a constant probability $p_\text{switch}$ that switches to another environment. For
all experiments, we set $p_\text{switch}$ to 0.2. We truncate the maximum number of appearances
per class to 6. If the maximum appearance is reached, we will sample another class.

\subsection{Metrics}
\label{sec:metrics}
\paragraph{Average precision:} We chose to use AP (average precision or area under the precision-recall curve) as a way of integrating two aspects of performance:

\begin{enumerate}
    \item the binary accuracy of whether an instance belongs to a known or unknown class (KU-Assign for short), and
    \item the accuracy of assigning an instance the correct class label given it is from a known class (Class-Assign for short).
\end{enumerate}
The procedure to calculate AP is as follows. We first sort all the {KU-Assign, Class-Assign} predictions across all sequences in descending order based on KU-Assign probability, where the high ranked predictions should be known (not novel) classes. For the N top ranked instances in the sorted list, we compute:
\begin{enumerate}
    \item precision@N = correct(Class-Assign)@N / N
    \item recall@N = correct(Class-Assign)@N / K,
\end{enumerate}
where K is the true number of known instances and correct(Class-Assign)@N is the count of the number of correct class assignments among the top N. (The class assignment for an unknown instance is always incorrect.) To obtain the AP, we compute the integral of the function (y=precision@N, x=recall@N) across all N’s.

\paragraph{N-shot accuracy:} We define N-shot accuracy as the number of times an instance that has been seen N times thus far in the sequence is classified correctly. We compute the mean and standard error of this over all sequences.

\subsection{Additional \ourroom{} Statistics} 
Statistics of the \ourroom{} are included in Table~\ref{tab:dataset_stats}, in comparison to other
few-shot and continual learning datasets. Note that since \ourroom{} is collected from a simulated
environment, with 90 indoor worlds consisting of 1.2K panorama images and 1.22M video frames. The
dataset contains about 6.9K random walk sequences with a maximum of 200 frames per sequence. For
training we randomly crop 100 frames to form a training sequence. There are 7.0K unique instance
classes.

\begin{figure}[t]
\vspace{-0.2in}
\centering
\includegraphics[width=0.9\textwidth,trim={0cm 7cm 10.2cm 0cm},clip]{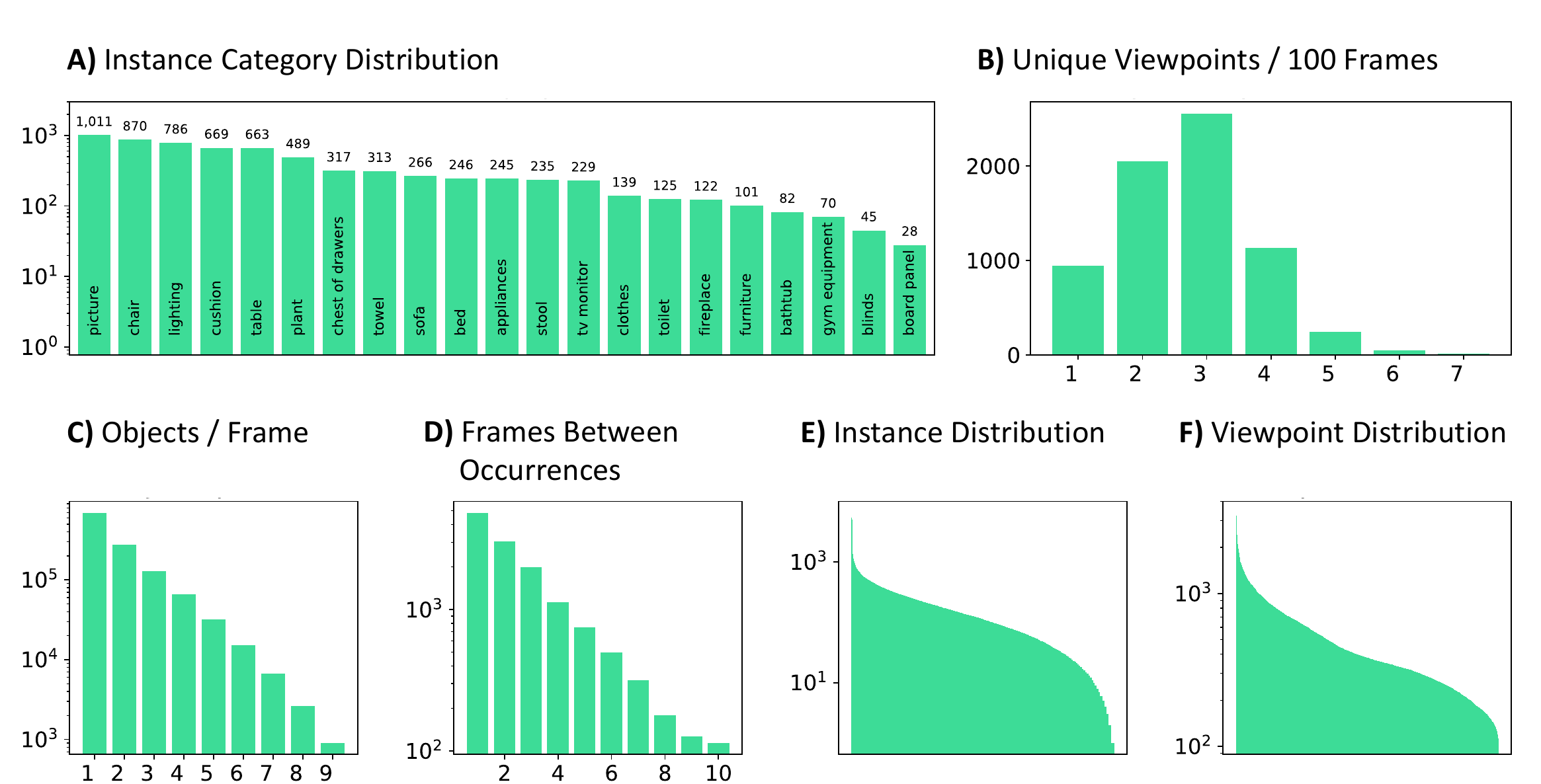}
\caption{Additional statistics about our \ourroom{} dataset.}
\label{fig:additionalstats}
\end{figure}

Plots of additional statistics of \ourroom{} are shown in Figure~\ref{fig:additionalstats}. In
addition to the ones shown in the main paper, instances and viewpoints also follow long tail
distributions. The number of objects in each frame follows an exponential distribution.

\begin{table}[t]
\vspace{-0.5in}
\begin{center}
\begin{small}
\caption{\small Continual \& few-shot learning datasets}
\label{tab:dataset_stats}
\begin{tabular}{ccccccc}
\toprule
                              & {\bf Images} & {\bf Sequences} & {\bf Classes} & {\bf Content}                  \\
\midrule                                                                     
Permuted MNIST~\citep{mnist}   & 60K          & -           & -          & Hand written digits          \\
Omniglot~\citep{omniglot}      & 32.4K        & -           & 1.6K       & Hand written characters      \\
CIFAR-100~\citep{cifar}        & 50K          & -           & 100        & Common objects               \\
mini-ImageNet~\citep{matchingnet}& 50K        & -           & 100        & Common objects               \\
tiered-ImageNet~\citep{fewshotssl}& 779K      & -           & 608        & Common objects               \\
OpenLORIS  \citep{openloris}   & 98K          & -           & 69         & Small table-top obj.         \\
CORe50  \citep{core50}         & 164.8K       & 11          & 50         & Hand-held obj.               \\
\midrule                                                                                                          
\ourroom{} (Ours)            & 1.22M         & 6.9K        & 7.0K       & General indoor instances     \\
\bottomrule
\end{tabular}
\vspace{-0.2in}
\end{small}
\end{center}
\end{table}

\subsection{\ourroom{} Simulator Details}
We generate our episodes with a two-stage process using two simulators -- HabitatSim~\citep{habitat}
and MatterSim~\citep{mattersim} -- because HabitatSim is based on 3D meshes and using HabitatSim
alone will result in poor image quality due to incorrect mesh reconstruction. Therefore we
sacrificed the continuous movement of agents within HabitatSim and base our environment navigation
on the discrete viewpoints in MatterSim, which is based on real panoramic images. The horizontal
field of view is set to 90 degrees for HabitatSim and 100 degrees for MatterSim, and we simulate
with\ 800$\times$600 resolution.

The first stage of generation involves randomly picking a sequence of viewpoints on the connectivity
graph within MatterSim. For each viewpoint, the agent scans the environment along the yaw and pitch
axes for a random period of time until a navigable viewpoint is within view. The time spent in a
single viewpoint follows a Gaussian distribution with mean 5.0 and standard deviation 1.0. At the
start of each new viewpoint, the agent randomly picks a direction to rotate and takes 12.5 degree
steps along the yaw axis, and with 95\% probability, a 5 degree rotation along the pitch axis is
applied in a randomly chosen direction. When a navigable viewpoint is detected, the agent will
navigate to the new viewpoint and reset the direction of scan. When multiple navigable viewpoints
are present, the agent uniformly samples one.

In the second stage, an agent in HabitatSim retraces the viewpoint path and movements of the first
stage generated by MatterSim, collecting mesh-rendered RGB and instance segmentation sensor data.
The MatterSim RGB and HabitatSim RGB images are then aligned via FLANN-based feature matching
~\citep{muja2009flann}, resulting in an alignment matrix that is used to place the MatterSim RGB and
HabitatSim instance segmentation maps into alignment. The sequence of these MatterSim RGB and
HabitatSim instance segmentation maps constitute an episode.

We keep objects of the following categories: \texttt{picture, chair, lighting, cushion, table,
plant, chest of drawers, towel, sofa, bed, appliances, stool, tv monitor, clothes, toilet,
fireplace, furniture, bathtub, gym equipment, blinds, board panel}. We initially generate 600 frames
per sequence and remove all the frames with no object. Then we store every 200 image frames into a
separate file.

During training and evaluation, each video sequence is loaded, and for each image we go through each
object present in the image. We create the attention map using the segmentation groundtruth of the
selected object. The attention map and the image together form a \textit{frame} in our model input.
For training, we randomly crop 100 frames from the sequence, and for evaluation we use the first 100
frames for deterministic results.

Please visit our released code repository to download the \ourroom{} dataset.

\begin{table*}[t]
\vspace{-0.5in}
\caption{Comparison of past FSL and CL paradigms vs. our online contextualized FSL (OC-FSL).}
\begin{center}
\begin{small}
\resizebox{\textwidth}{!}{
\begin{tabular}{ccccccc}
\toprule
\mr{2}{\tb{Tasks}}                    & \tb{Few}  & \tb{Semi-sup. } & \mr{2}{\tb{Continual}} & \tb{Online} & \tb{Predict}& \tb{Soft Context}  \\
                                      & \tb{Shot} & \tb{Supp. Set}  &                        & \tb{Eval.}  & \tb{New}    & \tb{Switch}        \\
\midrule                                                                                                                                                                          
Incremental Learning (IL) \citep{icarl}& \xm       & \xm             & \cm                    & \hx         & \xm         & \xm                \\
Few-shot (FSL) \citep{matchingnet}     & \cm       & \xm             & \xm                    & \xm         & \xm         & \xm                \\
Incremental FSL \citep{attnattractor}  & \cm       & \xm             & \hx                    & \xm         & \xm         & \xm                \\
Cls. Incremental FSL \citep{fscil}     & \cm       & \xm             & \cm                    & \hx         & \xm         & \xm                \\
Semi-supv. FSL \citep{fewshotssl}      & \cm       & \cm             & \xm                    & \xm         & \cm         & \xm                \\
MOCA \citep{moca}                      & \cm       & \xm             & \cm                    & \xm         & \xm         & \hx                \\
Online Mixture \citep{onlinemixture}   & \cm       & \xm             & \cm                    & \xm         & \xm         & \hx                \\
Online Meta \citep{oml}                & \cm       & \xm             & \cm                    & \xm         & \xm         & \xm                \\
Continual FSL* \citep{contfsl}         & \cm       & \xm             & \cm                    & \xm         & \xm         & \xm                \\
OSAKA* \citep{osaka}                   & \cm       & \xm             & \cm                    & \cm         & \hx         & \cm                \\
\midrule                                                                                                                                                                                         
OC-FSL (Ours)                         & \cm       & \cm             & \cm                    & \cm         & \cm         & \cm                \\
\bottomrule
\end{tabular}
}
\label{tab:benchmark}
\\
\vspace{0.05in}
* denotes concurrent work.
\end{small}
\end{center}
\end{table*}
\subsection{Semi-supervised Labels:}
Here we describe how we sample the labeled vs. unlabeled flag for each example in the
semi-supervised sequences in both \ourchar{} and \ourroom{} datasets. Due to the imbalance in our
class distribution (from both the Chinese restaurant process and real data collection), directly
masking the label may bias the model to ignore the rare seen classes. Ideally, we would like to
preserve at least one labeled example for each class. Therefore, we designed the following
procedure.

First, for each class $k$, suppose $m_k$ is the number of examples in the sequence that belong to
the class. Let $\alpha$ be the target label ratio. Then the class-specific label ratio $\alpha_k$
is:
\begin{align}
\alpha_k = (1 - \alpha) \exp(-0.5 (m_k - 1)) + \alpha.
\label{eq:semisup}
\end{align}
We then for each class $k$, we sample a binary Bernoulli sequence based on $\Ber(\alpha_k)$. If a
class has all zeros in the Bernoulli sequence, we flip the flag of one of the instances to 1 to make
sure there is at least one labeled instance for each class.
For all experiments, we set $\alpha = 0.3$.

\subsection{Dataset Splits}
We include details about our dataset splits in Table~\ref{tab:omniglotsplit} and
\ref{tab:matterportsplit}.

\section{Experiment Details}
\label{app:exp}
\subsection{Network Architecture}
For the \ourchar{} experiment we used the common 4-layer CNN for few-shot learning with 64 channels
in each layer, resulting in a 64-d feature vector~\citep{protonet}. For the \ourroom{} experiment we
resize the input to 120$\times$160 and we use the ResNet-12 architecture~\citep{tadam} with
\{32,64,128,256\} channels per block. To represent the feature of the input image with an attention
mask, we concatenate the global average pooled feature with the attention ROI feature, resulting in
a 512d feature vector. For the contextual RNN, in both experiments we used an LSTM~\citep{lstm} with
a 256d hidden state. 

We use a linear layer to map from the output of the RNN to the features and control variables. We
obtain $\gamma^{r,w}$ by adding 1.0 to the linear layer output and then applying the softplus
activation. The bias units for $\beta^{r,w}$ are initialized to 10.0. We
also apply the softplus activation to $\bm$ from the linear layer output.

\begin{table}[t]
\vspace{-0.5in}
\caption{\textbf{Split information for {\it \ourchar{}}}. Each column is an alphabet and we include all the characters in the alphabet in the split. Rows are continuation of lines.}
\begin{center}
\begin{small}
\label{tab:omniglotsplit}
\begin{tabular}{clll}
\toprule

\mr{11}{Train} &
\texttt{Angelic} &
\texttt{Grantha} &
\texttt{N Ko}\\
& 
\texttt{Aurek-Besh} &
\texttt{Japanese (hiragana)} &
\texttt{Malay}
\\
& 
\texttt{Asomtavruli} &
\texttt{Sanskrit} &
\texttt{Ojibwe}
\\
& 
\texttt{Korean} &
\texttt{Arcadian} &
\texttt{Greek}
\\
& 
\texttt{Alphabet of the Magi} &
\texttt{Blackfoot} &
\texttt{Futurama}
\\
& 
\texttt{Tagalog} &
\texttt{Anglo-Saxon Futhorc} &
\texttt{Braille}
\\
& 
\texttt{Cyrillic} &
\texttt{Burmese} &
\texttt{Avesta}
\\
& 
\texttt{Gujarati} &
\texttt{Ge ez} &
\texttt{Syriac (Estrangelo)}
\\
& 
\texttt{Atlantean} &
\texttt{Japanese (katakana)} &
\texttt{Balinese}
\\
& 
\texttt{Atemayar Qelisayer} &
\texttt{Glagolitic} &
\texttt{Tifinagh}
\\
& 
\texttt{Latin} &
\texttt{Inuktitut} &
\\
\midrule
\mr{2}{Val} &
\texttt{Hebrew} &
\texttt{Mkhedruli} &
\texttt{Armenian}\\
& 
\texttt{Early Aramaic} &
\texttt{Bengali} &
\\
\midrule

\mr{5}{Test} & 
\texttt{Gurmukhi} &
\texttt{Kannada} & 
\texttt{Keble} \\
&
\texttt{Malayalam} &
\texttt{Manipuri} &
\texttt{Mongolian} 
\\
&
\texttt{Old Church Slavonic} &
\texttt{Oriya} &
\texttt{Syriac (Serto)} \\
&
\texttt{Sylheti} &
\texttt{Tengwar} &
\texttt{Tibetan}\\
&
\texttt{ULOG}
\\
\bottomrule
\end{tabular}
\end{small}
\end{center}
\end{table}
\begin{table}[t]
\caption{\textbf{Split information for {\it \ourroom{}}}. Each column is the ID of an indoor world. Rows are continuation of the lines.}
\label{tab:matterportsplit}
\begin{center}
\begin{small}
\begin{tabular}{cc}
\toprule
\mr{12}{Train} &
\texttt{
r1Q1Z4BcV1o
JmbYfDe2QKZ
29hnd4uzFmX
ULsKaCPVFJR
E9uDoFAP3SH
}\\
&
\texttt{
8WUmhLawc2A
Uxmj2M2itWa
mJXqzFtmKg4
V2XKFyX4ASd
EU6Fwq7SyZv
}\\
&
\texttt{
gYvKGZ5eRqb
gxdoqLR6rwA
YFuZgdQ5vWj
gTV8FGcVJC9
sT4fr6TAbpF
}\\
&
\texttt{
VVfe2KiqLaN
fzynW3qQPVF
WYY7iVyf5p8
VFuaQ6m2Qom
YmJkqBEsHnH
}\\
&
\texttt{
2t7WUuJeko7
pLe4wQe7qrG
cV4RVeZvu5T
XcA2TqTSSAj
ur6pFq6Qu1A
}\\
&
\texttt{
1pXnuDYAj8r
b8cTxDM8gDG
x8F5xyUWy9e
X7HyMhZNoso
aayBHfsNo7d
}\\
&
\texttt{
TbHJrupSAjP
sKLMLpTHeUy
2azQ1b91cZZ
2n8kARJN3HM
Vvot9Ly1tCj
}\\
&
\texttt{
S9hNv5qa7GM
EDJbREhghzL
qoiz87JEwZ2
q9vSo1VnCiC
Vt2qJdWjCF2
}\\
&
\texttt{
VzqfbhrpDEA
D7G3Y4RVNrH
ZMojNkEp431
uNb9QFRL6hY
5LpN3gDmAk7
}\\
&
\texttt{
rqfALeAoiTq
e9zR4mvMWw7
yqstnuAEVhm
zsNo4HB9uLZ
JF19kD82Mey
}\\
&
\texttt{
759xd9YjKW5
wc2JMjhGNzB
rPc6DW4iMge
jh4fc5c5qoQ
HxpKQynjfin
}\\
&
\texttt{
GdvgFV5R1Z5
kEZ7cmS4wCh
vyrNrziPKCB
D7N2EKCX4Sj
PX4nDJXEHrG
}\\

\midrule

\mr{2}{Val} &
\texttt{
s8pcmisQ38h
dhjEzFoUFzH
RPmz2sHmrrY
1LXtFkjw3qL
8194nk5LbLH
}\\
&
\texttt{
jtcxE69GiFV
QUCTc6BB5sX
p5wJjkQkbXX
JeFG25nYj2p
82sE5b5pLXE
}\\
\midrule

\mr{4}{Test} & 
\texttt{
oLBMNvg9in8
r47D5H71a5s
Z6MFQCViBuw
YVUC4YcDtcY
pRbA3pwrgk9
}\\
&
\texttt{
SN83YJsR3w2
gZ6f7yhEvPG
ac26ZMwG7aT
7y3sRwLe3Va
B6ByNegPMKs
}\\
&
\texttt{
UwV83HsGsw3
VLzqgDo317F
17DRP5sb8fy
pa4otMbVnkk
5ZKStnWn8Zo
}\\
&
\texttt{
PuKPg4mmafe
Pm6F8kyY3z2
i5noydFURQK
ARNzJeq3xxb
5q7pvUzZiYa
}\\
\bottomrule
\end{tabular}
\end{small}
\end{center}
\end{table}

\begin{table}[t]
\vspace{-0.5in}
\caption{{\it \ourroom{}} dataset split size}
\label{tab:matterportsplitsize}
\begin{center}
\begin{small}
\begin{tabular}{cccc}
\toprule
Split & Worlds & Sequences & Frames \\
\midrule
Train & 60     & 4,699      &   823,444     \\
Val   & 20     &  725         &  125,823  \\
Test  & 10     &  1,547      &  271,335      \\
\midrule
Total & 90     & 6,971   & 1,220,602 \\
\bottomrule
\end{tabular}
\end{small}
\end{center}
\end{table}

\subsection{Training Procedure}
We use the Adam optimizer~\citep{adam} for all of our experiments, with a gradient cap of 5.0. For
\ourchar{} we train the network for 40k steps with a batch size 32 and maximum sequence length 150
across 2 GPUs and an initial learning rate 2e-3 decayed by 0.1$\times$ at 20k and 30k steps. For
\ourroom{} we train for 20k steps with a batch size 8 and maximum sequence length 100 across 4 GPUs
and an initial learning rate 1e-3 decayed by 0.1$\times$ at 8k and 16k steps. We use the BCE coefficient
$\lambda=1$  for all experiments. In semi-supervised experiments, around 30\% examples are labeled when the number of examples grows large ($\alpha = 0.3$, see Equation~\ref{eq:semisup}). Early stopping is used in \ourroom{} experiments
where the checkpoint with the highest validation AP score is chosen.
For \ourroom{}, we sample Bernoulli sequences on unlabeled inputs to 
gradually allow semi-supervised writing to the prototype memory and we find it helps training stability. The probability starts with 0.0 and increase by 0.2 every 2000 training steps until reaching 1.0.

\subsection{Data Augmentation}
For \ourchar{}, we pad the 28$\times$28 image to 32$\times$32 and then apply random cropping.

For \ourroom{}, we apply random cropping in the time dimension to get a chunk of 100 frames per
input example. We also apply random dropping of 5\% of the frames. We pad the 120$\times$160 images
to 126 $\times$ 168 and apply random cropping in each image frame. We also randomly flip the order
of the sequence (going forward or backward).

\subsection{Spatiotemporal context experiment details}
We use the Kylberg texture dataset~\citep{uppsala} without rotations. Texture classes are split into train, val, and test, defined in Table~\ref{tab:uppsalasplit}. We resize all images first to 256$\times$256. For each Omniglot image, a 28$\times$28 patch is randomly cropped from a texture image to serve as background. Random Gaussian noises with mean zero and standard deviation 0.1 are added to the background images.

For spatial background experiments, we added an additional learnable network of the same size as the main network to take the background image as input, and output the same sized embedding vector. This embedding vector is further concatenated with the main embedding vector to form the final embedding of the input. We also found that using spatially overlayed images with a single CNN can achieve similar performance as well. The final numbers are reported using the concatenation approach since it is less prone to overlay noises and is more similar to the implementation we use in the RoamingRooms experiments.

\begin{table}[t]
\vspace{-0.5in}
\caption{\textbf{Split information for the Kylberg texture dataset}. Each column is an texture type. Rows are continuation of lines.}
\label{tab:uppsalasplit}
\begin{center}
\begin{small}
\begin{tabular}{clllll}
\toprule

\mr{3}{Train} &
\texttt{blanket2} &
\texttt{ceiling2} &
\texttt{floor2} &
\texttt{grass1} &
\texttt{linseeds1} \\
&
\texttt{pearlsugar1} &
\texttt{rice2} &
\texttt{scarf2} &
\texttt{screen1} &
\texttt{seat2} \\
&
\texttt{sesameseeds1} &
\texttt{stone1} & 
\texttt{stoneslab1} & &
\\
\midrule

\mr{2}{Val} &
\texttt{blanket1} &
\texttt{canvas1} &
\texttt{ceiling1} & 
\texttt{floor1} &
\texttt{scarf1} \\
&
\texttt{rice1} &
\texttt{stone2} & & & \\
\midrule
\mr{2}{Test} & 
\texttt{wall1} &
\texttt{lentils1} & 
\texttt{cushion1} &
\texttt{rug1} &
\texttt{sand1} \\
&
\texttt{oatmeal1} &
\texttt{stone3} &
\texttt{seat1} & & \\
\bottomrule
\end{tabular}
\end{small}
\end{center}
\end{table}

\subsection{Baseline implementation details}
\paragraph{Online meta-learning (OML):} The OML  model performs one gradient descent step for each
input. In order for the model to predict unknown, we use the probability output from the softmax
layer summing across the unused units. For example, if the softmax layer has 40 units and we have
only seen 5 classes so far, then we sum the probability from the 6th to the last units. This summed
probability is separately trained with a binary cross entropy, same as in Equation~\ref{eq:loss}.

The inner learning rate is set to 1e-2 and we truncate the number of unrolled gradient descent steps
to 5/20 (\ourchar{}/\ourroom{}), in order to make the computation feasible. For \ourchar{}, the
network is trained with a batch size 32 across 2 GPUs, for a total of 20k steps, with an initial
learning rate 2e-3 decayed by 0.1 at 10k and 16.7k steps. For \ourroom{}, the network is trained
with a batch size 8 across 4 GPUs, for a total of 16k steps, with an initial learning rate 1e-3
decayed by 0.1 at 6.4k and 12.8k steps.

\paragraph{Long short-term memory (LSTM):} We apply a stacked two layer LSTM with 256 hidden
dimensions. Inputs are $\bh_t^{\text{CNN}}$ concatenated with the label one-hot vector. If an
example is unlabeled, then the label vector is all-zero. We directly apply a linear layer on top of
the LSTM to map the LSTM memory output into classification logits, and the last logit is the binary
classification logit reserved for unknown. The training procedure is the same as our CPM model.

\paragraph{Differentiable neural computer (DNC):} In order to make the DNC model work properly, we
found that it is sometimes helpful to pretrain the CNN weights. Simply initializing from scratch and
train CNN+DNC end-to-end sometimes results in poor performance. We hypothesize that the attention
structure in the DNC model is detrimental to representation learning. Therefore, for \ourchar{}
experiments, we use pretrained ProtoNet weights for solving 1-shot 5-way episodes to initialize the
CNN, and we keep finetuning the CNN weights with 10\% of the full learning rate. For \ourroom{}
experiments, we train the full model end-to-end from scratch.

The DNC is also modified so that it is more effective using the label information from the input. In
the original MANN paper~\citep{mann} for one-shot learning, the input features $\bh_t^{\text{CNN}}$
and the label one-hot ID are simply concatenated to feed into the LSTM controller of MANN. We find
that it is beneficial to directly add label one-hot vector as an input to the write head that
generates the write attention and the write content.  Similar to the LSTM model, the memory readout is also sent to a
linear layer in order to get the final classification logits, and the last logit is the binary
classification logit reserved for the unknowns. Finally we remove the linkage prediction part of the DNC
due to training instability.

The controller LSTM has 256 hidden dimensions, and the memory has 64 slots each with 64 dimensions.
There are 4 read heads and 4 write heads. The training procedure is the same as CPM.

\paragraph{Online ProtoNet:} Online ProtoNet is our modification of the original
ProtoNet~\citep{protonet}. It is similar to our CPM model without the contextual RNN. The feature
from the CNN is directly written to the prototype memory. In addition, we do not predict the control
hyperparameters $\beta^{\{r,w\}}_t,\gamma^{\{r,w\}}_t$ from the RNN and they are learned as regular parameters. The training procedure is the same as CPM.

\paragraph{Online MatchingNet:} Online MatchingNet is our modification of the original
MatchingNet~\citep{matchingnet}. We do not consider the context embedding in the MatchingNet paper
since it was originally designed for the entire episode using an attentional RNN encoder. It is
similar to online ProtoNet but instead of doing online averaging, it directly stores each example
and its class. Since it is an example-based storage, we did not extend it to learn from
unlabeled examples, and all unlabeled examples are skipped. We use a similar decision rule to
determine whether an example belongs to a known cluster by looking at the distance to the nearest
exemplar stored in the memory, shifted by $\beta$ and scaled by $1/\gamma$. Note that online
MatchingNet is not efficient at memory storage since it scales with the number of steps in the
sequence. In addition, we use the negative Euclidean distance as the similarity function. The training
procedure is the same as CPM.

\paragraph{Online infinite mixture prototypes (IMP):} Online IMP is proposed as a mix of prototype and example-based storage by allowing a class to have multiple clusters. If an
example is classified as unknown or it is unlabeled, we will assign its cluster based on our
prediction, which either assigns it to one of the existing clusters or creates a new cluster,
depending on its distance to the nearest cluster. If a cluster with an unknown label later is
assigned with an example with a known class, then the cluster label will also be updated. We use the same
decision rule as online ProtoNet to determine whether an example belongs to a known cluster by
looking at the distance to the nearest cluster, shifted by $\beta$ and scaled by $1/\gamma$. As
described above, online IMP has the capability of learning from unlabeled examples, unlike online
MatchingNet. However similar to online MatchingNet, online IMP is also not efficient at memory
storage since in the worst case it also scales with the number of steps in the sequence. Again, the
training procedure is the same as CPM.

\section{Additional Experimental Results}
\label{sec:additionalresults}
\subsection{Effect of Forgetting}
We report the effect of forgetting of \ourroom{} and \ourimg{} in Table~\ref{tab:forgetroom} and \ref{tab:forgetimagenet}.
\begin{table}[t]
\vspace{-0.5in}
    \centering
    \caption{\textbf{Effect of forgetting over a time interval on \ourroom{}.} Average accuracy vs. the number of time steps since the model has last seen the label of a particular class.}
    
    \resizebox{\textwidth}{!}{
    \begin{tabular}{ccccccc|cccccc}
    \toprule
    & \mc{6}{c}{\bf Supervised} & \mc{6}{c}{\bf Semi-Supervised}\\
                &   1 - 2&      3 - 5&      6 - 10&     11 - 20&    21 - 50&    51 - 100&
                    1 - 2&      3 - 5&      6 - 10&     11 - 20&    21 - 50&    51 - 100\\
    \midrule
    OPN 1-Shot  &   93.5&   	89.3&	    79.4&	    67.2&	    60.3&	    60.1&
                    86.5&	    83.6&	    76.3&	    68.4&	    64.7&	    61.5\\
    CPM 1-Shot  &   \bf{95.7}&	\bf{92.2}&  \bf{85.7}&	\bf{75.2}&	\bf{70.0}&  \bf{66.4}&
                    \bf{91.0}&	\bf{88.7}&	\bf{82.9}&	\bf{77.0}&  \bf{72.2}&	\bf{66.5}\\
    \midrule
    OPN 3-Shot  &   95.1&	    91.8&	    85.6&   	78.2&	    74.6&	    73.8&
                    92.6&       88.0&	    85.1&	    81.1&	    80.6&	    76.7\\
    CPM 3-Shot  &   \bf{96.1}&	\bf{93.8}&	\bf{87.7}&	\bf{81.4}&  \bf{79.1}&	\bf{78.2}&
                    \bf{94.8}&  \bf{91.0}&	\bf{86.9}&	\bf{83.1}&	\bf{82.7}&  \bf{79.2}\\
    \bottomrule
    \end{tabular}}
    \label{tab:forgetroom}
\end{table}

\begin{table}[t]
\vspace{-0.2in}
    \centering
    \caption{\textbf{Effect of forgetting over a time interval on \ourimg{}.} Average accuracy vs. the number of time steps since the model has last seen the label of a particular class.}
    
    \resizebox{\textwidth}{!}{
    \begin{tabular}{ccccccc|cccccc}
    \toprule
    & \mc{6}{c}{\bf Supervised} & \mc{6}{c}{\bf Semi-Supervised}\\
                &   1 - 2&      3 - 5&      6 - 10&     11 - 20&    21 - 50&    51 - 100&
                    1 - 2&      3 - 5&      6 - 10&     11 - 20&    21 - 50&    51 - 100\\
    \midrule
    OPN 1-Shot  &   40.8&	35.9&	33.0&	\bf{30.7}&	\bf{27.0}&	\bf{21.4}&
                    40.5&	37.7&	35.9&	\bf{33.7}&	\bf{31.5}&	\bf{28.4}\\
    CPM 1-Shot  &   \bf{67.5}&	\bf{52.9}&	\bf{35.5}&	24.2&	18.3&	13.8&
                    \bf{60.4}&	\bf{51.3}&	\bf{39.5}&	26.6&	21.8&	15.4\\
    \midrule
    OPN 3-Shot  &   52.5&	50.3&	\bf{48.8}&	\bf{47.2}&	\bf{44.4}&	\bf{42.3}&
                    57.6&	55.1&	\bf{54.6}&	\bf{52.3}&	\bf{52.1}&	\bf{49.5}\\
    CPM 3-Shot  &   \bf{77.8}&	\bf{64.5}&	46.6&	32.9&	24.7&	17.9&
                    \bf{76.1}&	\bf{61.8}&	48.6&	30.5&	24.1&	15.6\\
    \bottomrule
    \end{tabular}}
    \label{tab:forgetimagenet}
\end{table}


\subsection{Embedding Visualization} 
\begin{figure}[t]
\centering
\begin{tabular}{cc}
Online ProtoNet & CPM (Ours) \\
\includegraphics[height=3.8cm, trim={2.5cm 1cm 2cm 1cm}, clip]{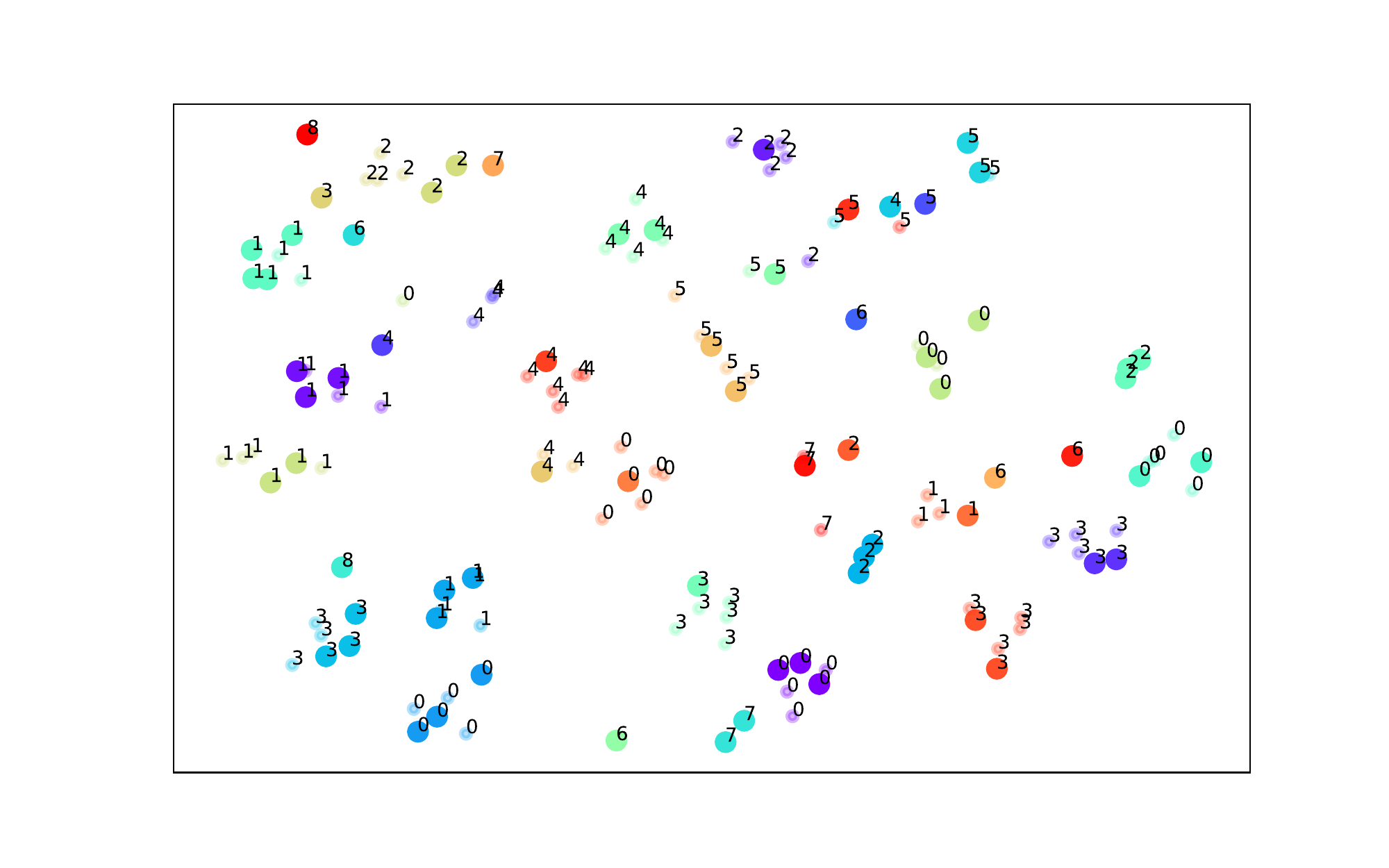}
&
\includegraphics[height=3.8cm, trim={2.5cm 1cm 2cm 1cm}, clip]{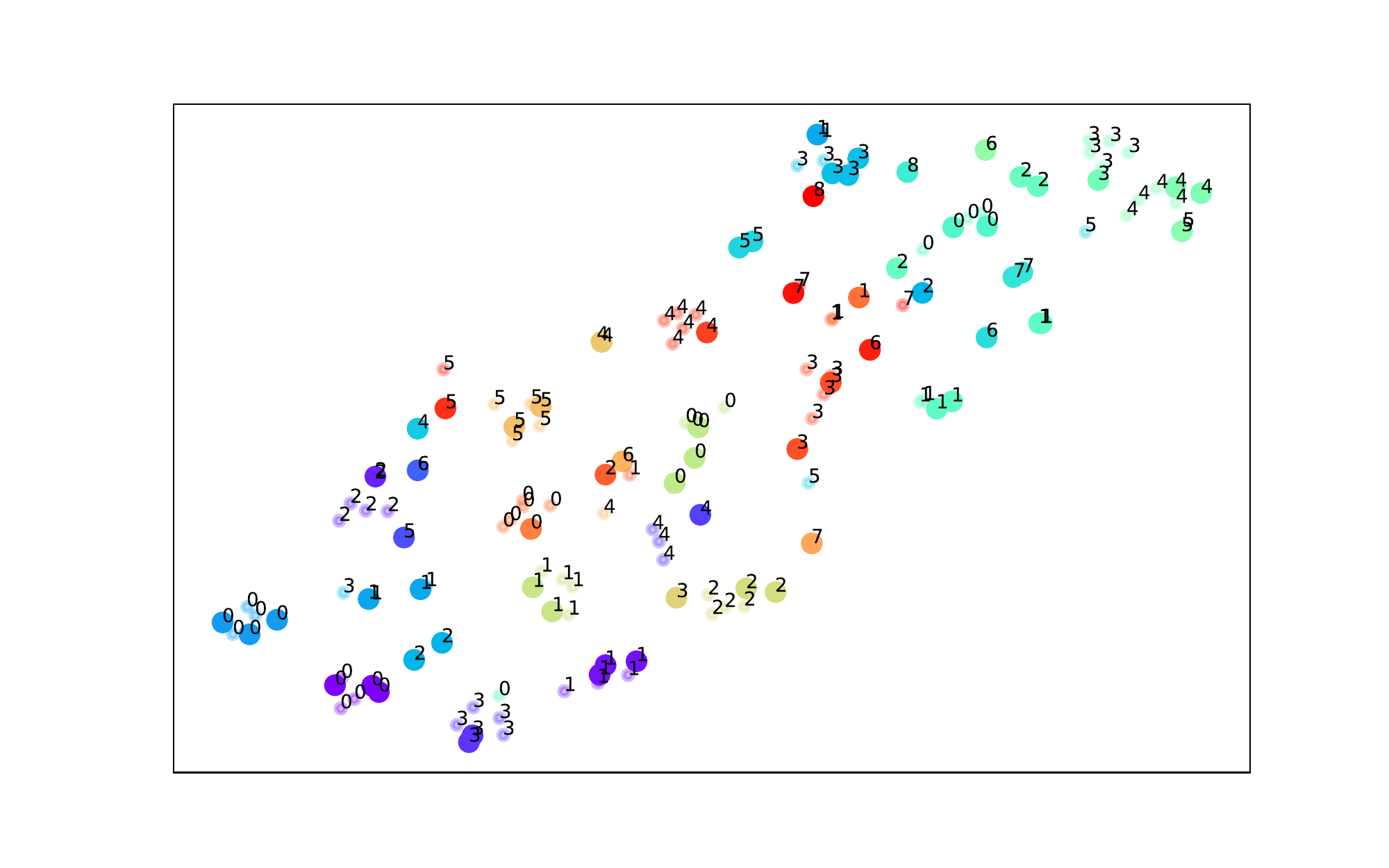}
\\
\includegraphics[height=3.8cm, trim={2.5cm 1cm 2cm 1cm}, clip]{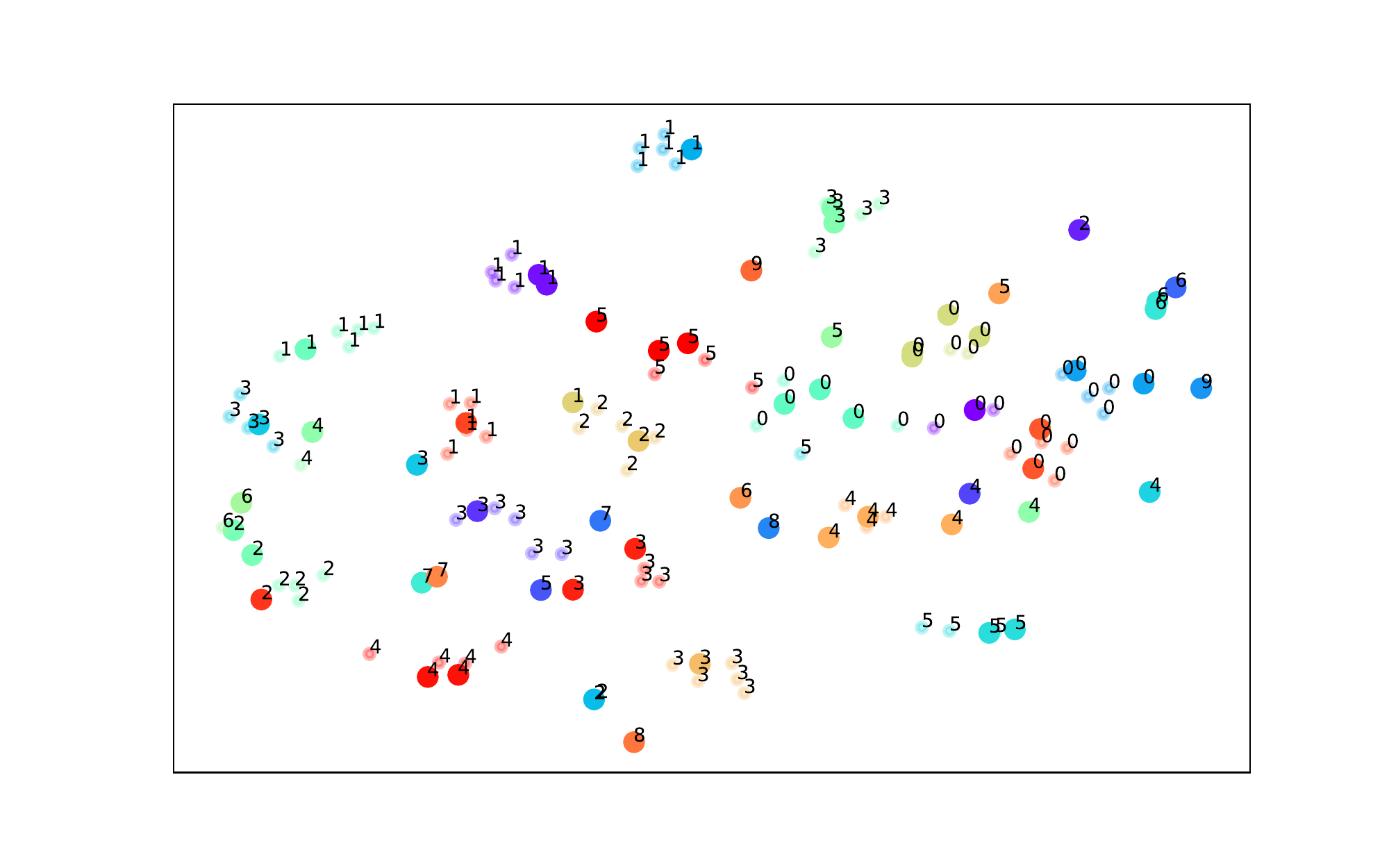}
&
\includegraphics[height=3.8cm, trim={2.5cm 1cm 2cm 1cm}, clip]{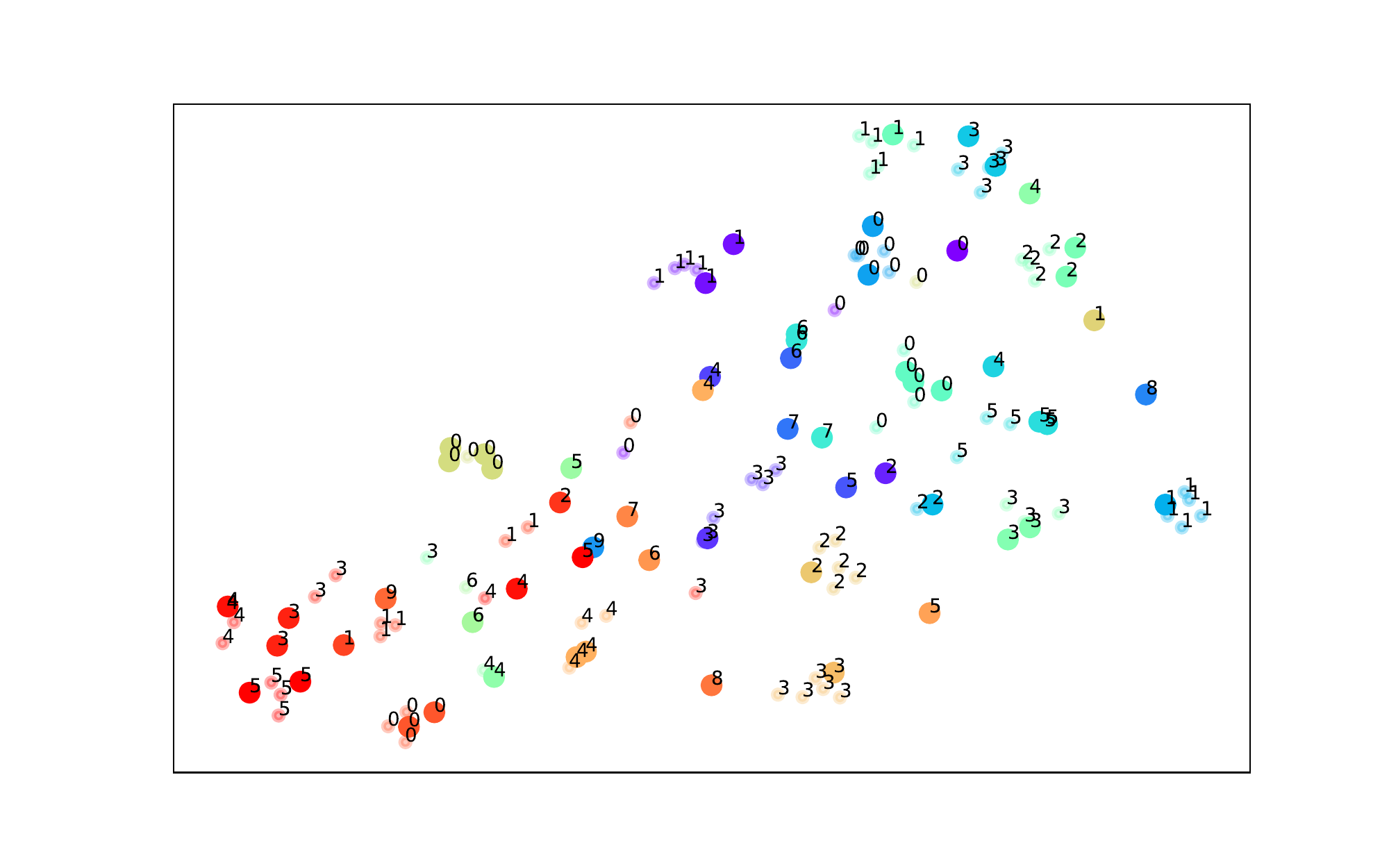}
\end{tabular}
\caption{\textbf{Embedding space visualization of \ourchar{} sequences using t-SNE~\citep{tsne}}. Different color
denotes different environments. Text labels (relative to each environment) are annotated beside the
scatter points. Unlabeled examples shown in smaller circles with lighter colors. \textbf{Left:}
Online ProtoNet; \textbf{Right:} CPM. The embeddings learned CPM model shows a smoother transition
of classes based on their temporal environments.}
\label{fig:tsne}
\end{figure}

Figure~\ref{fig:tsne} shows the learned embedding of each example in Online ProtoNet vs. our CPM
model in \ourchar{} sequences, where colors indicate environment IDs. In Online ProtoNet, the
example features does not reflect the temporal context, and as a result, colors are scattered across
the space. By contrast, in the CPM embedding visualization, colors are clustered together and we see
a smoother transition of environments in the embedding space.

\subsection{Control Parameters vs. Time}
\begin{figure}
\centering
\begin{tabular}{cc}
\includegraphics[height=4.0cm,trim={0.3cm 0cm 0.5cm 0},clip]{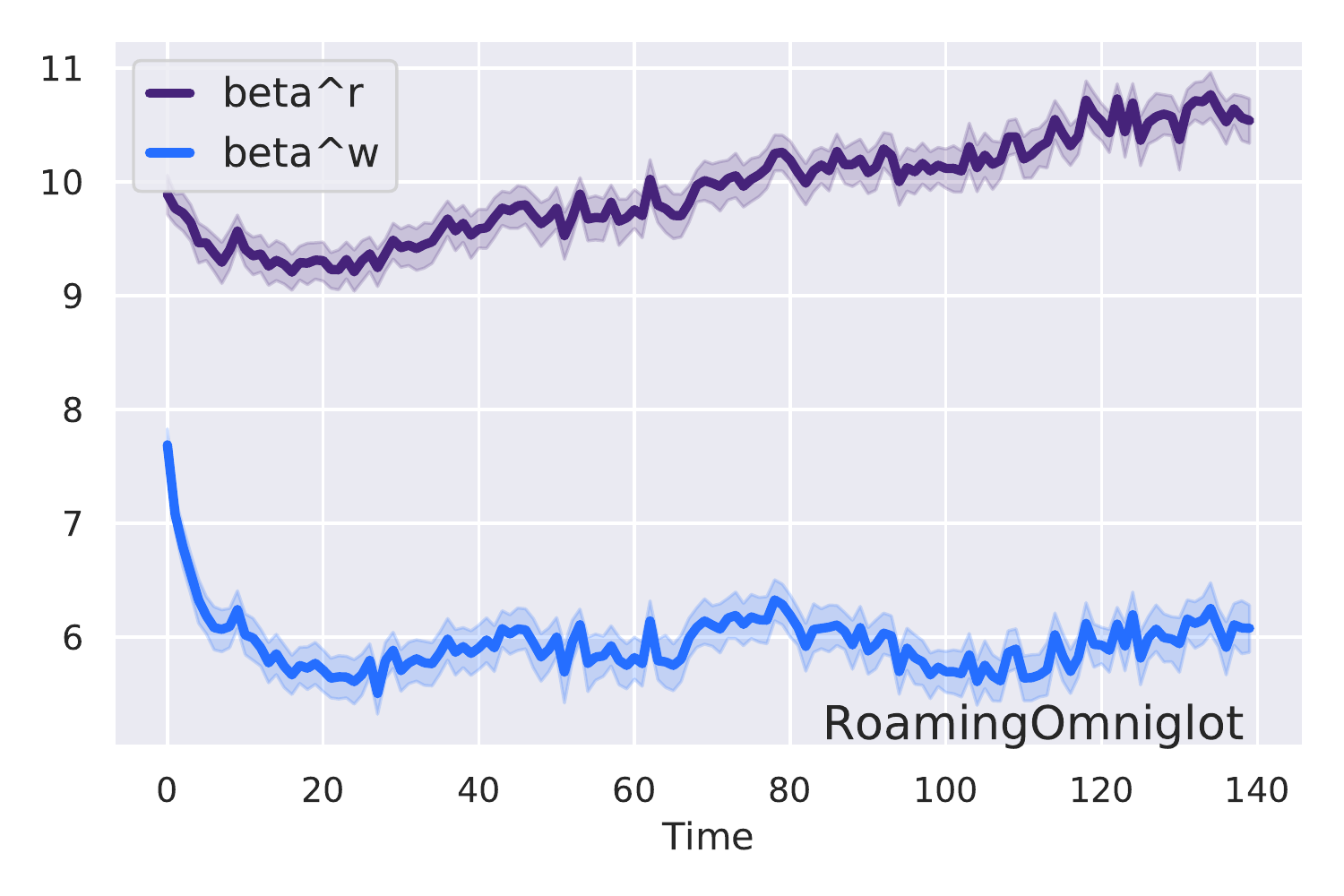}
\quad
&
\includegraphics[height=4.0cm,trim={0.3cm 0cm 0cm 0},clip]{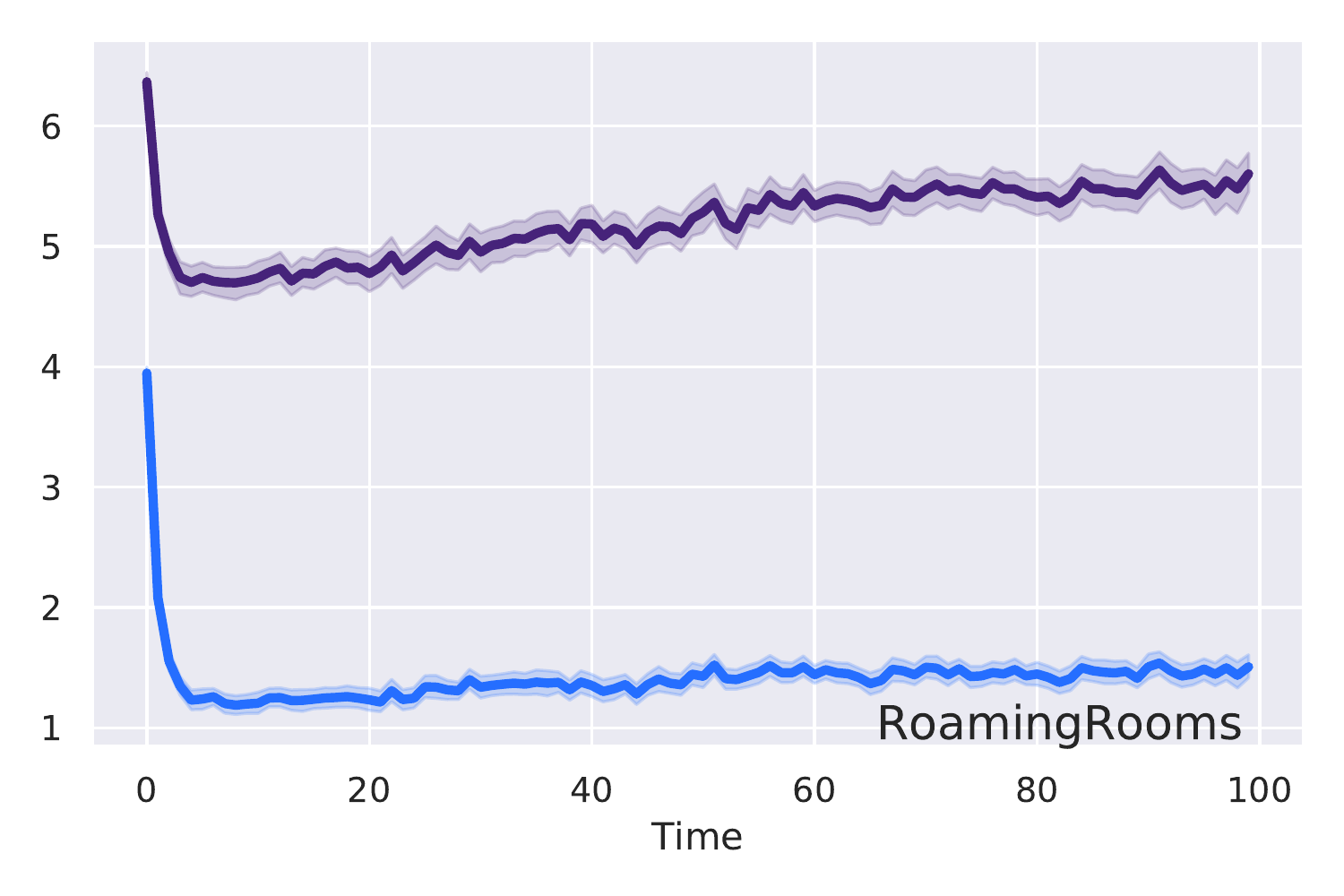}
\\
\includegraphics[height=4.0cm,trim={0.3cm 0cm 0.5cm 0},clip]{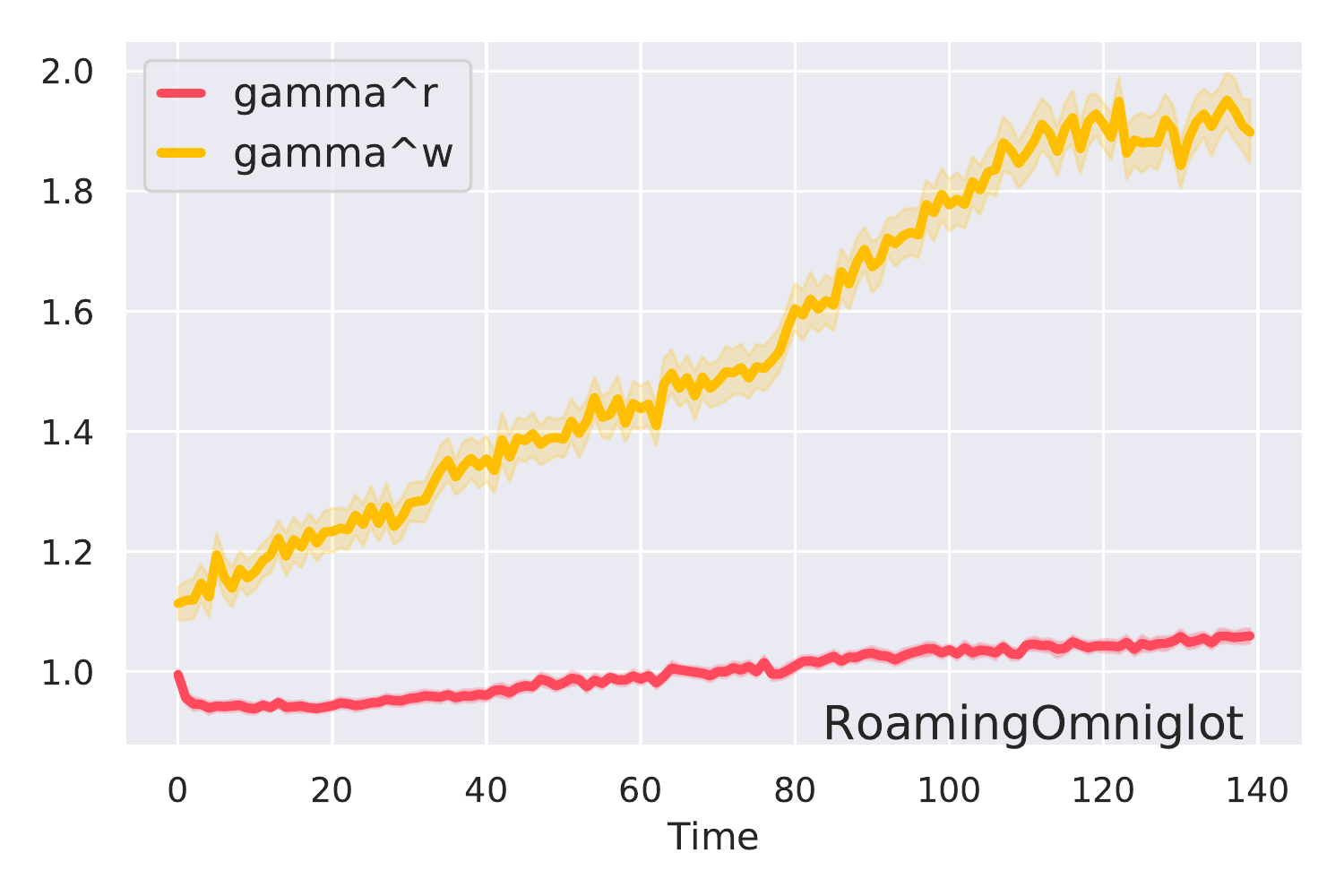}
\quad
&
\includegraphics[height=4.0cm,trim={0.3cm 0cm 0cm 0},clip]{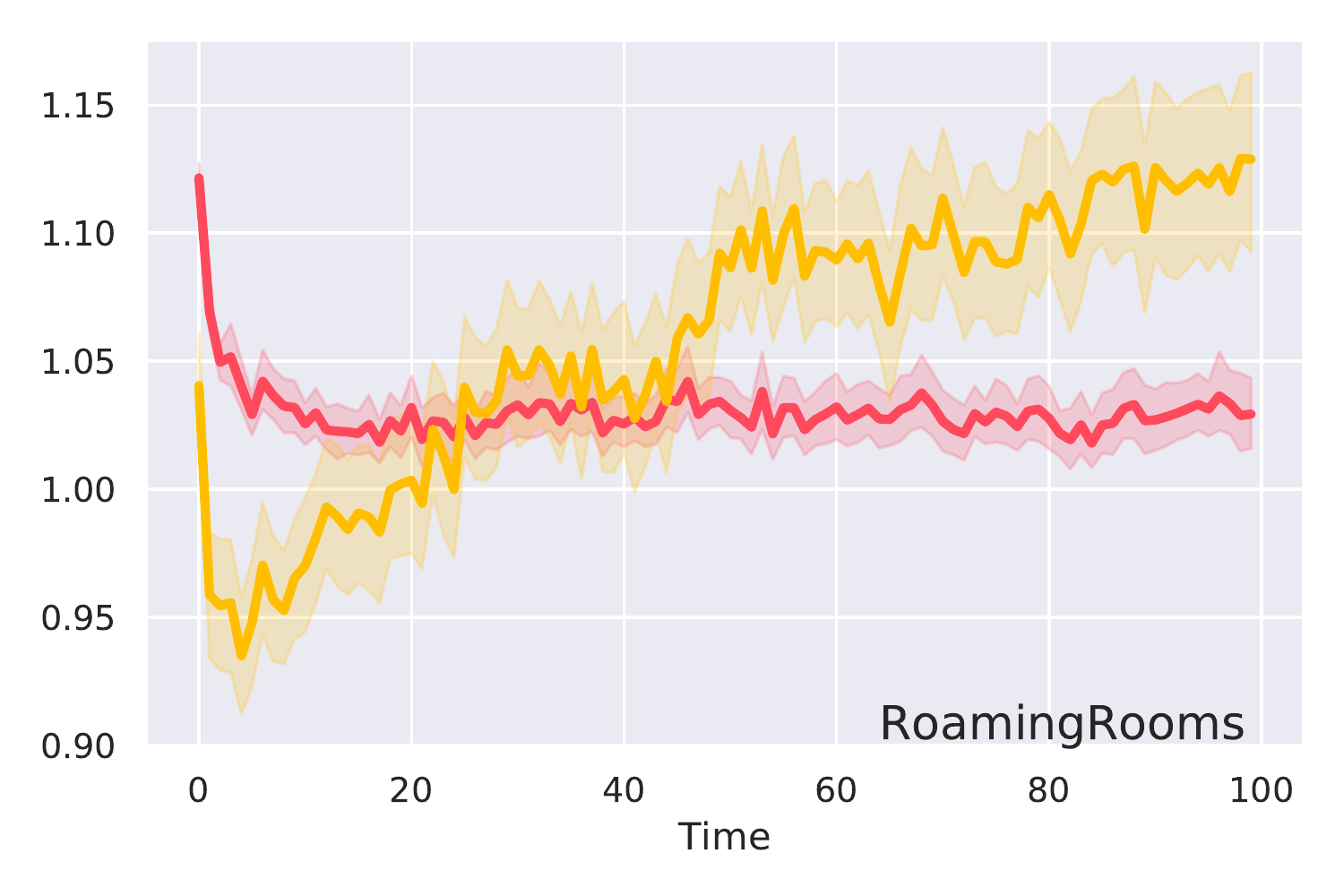}
\\
\end{tabular}
\vspace{-0.1in}
\caption{\textbf{CPM control parameters ($\beta^{r,w}, \gamma^{r,w}$) vs. time.}
\textbf{Left:} \ourchar{} sequences; \textbf{Right:} \ourroom{} sequences; \textbf{Top:}
$\beta^{r,w}$ the threshold parameter; \textbf{Bottom:} $\gamma^{r,w}$ the temperature parameter.}
\label{fig:betagamma}
\end{figure}

Finally we visualize the control parameter values predicted by the RNN in
Figure~\ref{fig:betagamma}. We verify that we indeed need two sets of $\beta$ and $\gamma$ for read
and write operations separately as they learn different values. $\beta^w$ is smaller than $\beta^r$
which means that the network is more conservative when writing to prototypes. $\gamma^w$ grows
larger over time, which means that the network prefers a softer slope when writing to prototypes
since in the later stage the prototype memory has already stored enough content and it can grow
faster, whereas in the earlier stage, the prototype memory is more conservative to avoid embedding
vectors to be assigned to wrong clusters.

\end{document}